%% file: main.tex
\pdfoutput=1

\documentclass[11pt]{article}

\usepackage[final]{acl}

\usepackage{times}
\usepackage{latexsym}
\usepackage[table]{xcolor}

\usepackage[T1]{fontenc}

\usepackage[utf8]{inputenc}
\usepackage{varwidth}

\usepackage{microtype}

\usepackage{inconsolata}

\usepackage{graphicx}
\usepackage{soul}
\usepackage{hyperref}
\usepackage{url}
\usepackage[font=small]{caption}
\usepackage{amsmath}
\usepackage{amsthm}
\usepackage{booktabs}
\usepackage{algorithm}
\usepackage{algorithmic}
\usepackage[switch]{lineno}
\usepackage{verbatim}

\usepackage{amssymb}%
\usepackage{pifont}%

\definecolor{bl}{RGB}{0,0,0}
\definecolor{gr}{RGB}{37,37,37}
\definecolor{sil}{RGB}{240,240,240}
\definecolor{gray}{RGB}{200,200,200}

\definecolor{green1}{HTML}{0b5400}
\definecolor{orange1}{HTML}{f3905c}
\definecolor{purple1}{HTML}{9258cc}
\definecolor{blue1}{HTML}{027db5}
\definecolor{pink1}{HTML}{ff7a7a}
\definecolor{darkgreen}{HTML}{005e19}
\definecolor{darkblue}{HTML}{240394}
\definecolor{darkred}{HTML}{C00000}
\definecolor{lightblue2}{HTML}{DEEBF7}
\definecolor{lightgreen2}{HTML}{E2F0D9}
\definecolor{lightgray2}{HTML}{767171}
\definecolor{lightgray}{HTML}{D3D3D3}

\title{STARLING: Self-supervised Training of Text-based Reinforcement Learning Agent with Large Language Models}

\author{Shreyas Basavatia\thanks{Equal Contributions}\thanks{Work done while SB was a high school student at Pelham Memorial High School. Code and data can be found at \url{https://github.com/IBM/starling-agent}.} \\
  Georgia Institute of Technology \\
  {\small \texttt{\url{sbasavatia3@gatech.edu}}} \\\And
  Keerthiram Murugesan\footnotemark[1] \\
  IBM Research \\
  \texttt{\small \url{keerthiram.murugesan@ibm.com}} \\\And
  Shivam Ratnakar\footnotemark[1] \\
  IBM Consulting \\
  \texttt{\small \url{shivam.ratnakar@st.niituniversity.in}}
  }

\begin{document}

\maketitle

\begin{abstract}
    Interactive fiction games have emerged as an important application to improve the generalization capabilities of language-based reinforcement learning (RL) agents. Existing environments for interactive fiction games are domain-specific or time-consuming to generate and do not train the RL agents to master a specific set of skills. In this work, we introduce an interactive environment for self-supervised RL, \textit{STARLING}, for text-based games that bootstraps the text-based RL agents with automatically generated games (based on the seed set of game ideas)  to boost the performance and generalization capabilities to reach a goal of the target environment. These games let the agent hone their skills on a predefined set of tasks. We create and test an environment with $100$ games, generated using this automated framework that uses large language models (GPT3) and an interactive fiction game engine (based on Inform7) to provide the user with the ability to generate more games under minimal human supervision. Experimental results based on both the human participants and baseline text-based RL agents reveal that current state-of-the-art text-based RL agents cannot use previously learned skills in new situations at the level humans can. These results enforce STARLING’s potential to serve as a sandbox environment for further research in self-supervised text-based RL.  
\end{abstract}

\section{Introduction}

\begin{figure}[t]
    \centering
    \includegraphics[scale=0.52]{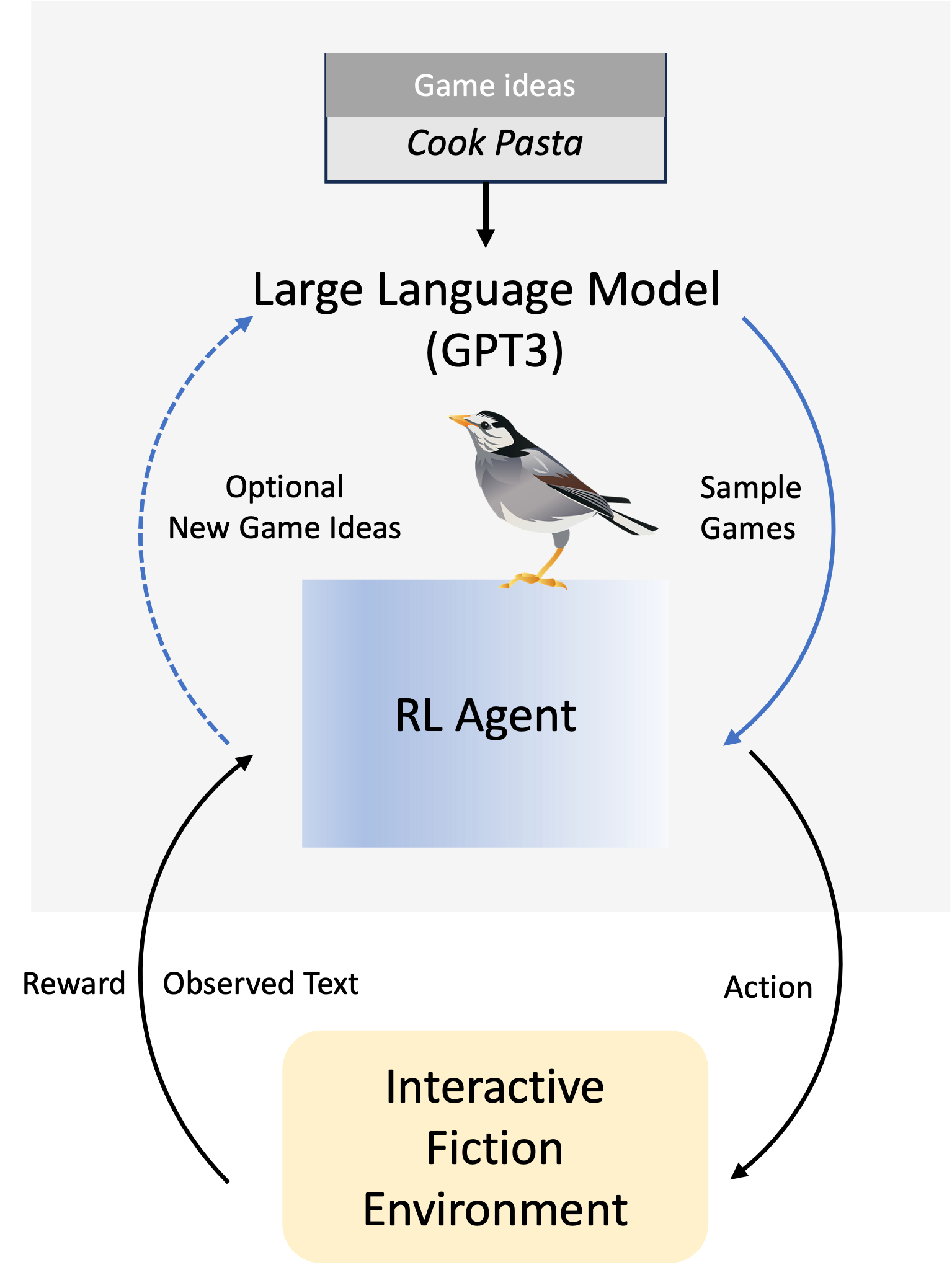}
    \caption{Architecture diagram for Self-supervised Text-based Reinforcement Learning using LLM (STARLING).}
    \label{fig:arch}
\end{figure}

Interactive fiction games such as Zork can be utilized as an important test-bed to improve the generalization capabilities of text-based reinforcement learning (TBRL) agents \cite{hausknecht2020interactive,jansen-2022-systematic}. In these games, both the observed state of the game and the actions taken are in natural language.  To play these games, the agents (or human players) need to understand the observed text from the environment and take relevant action toward the goal. These games encourage agents to understand the underlying state of the game and take actions to interact with the environment. In order to be successful, agents must use previously learned skills in new situations to complete an overarching goal. Current environments of interactive fiction games suffer from two major problems. First, environments such as TextWorld Commonsense measure simple commonsense reasoning based on one-hop relationships between entities (e.g., apple $\rightarrow$ refrigerator) \cite{murugesan2021text} but lack game complexity (besides a fewer number of games) to learn skills and generalize to novel domains.
Second, environments such as ScienceWorld (even though many variations of task-based games are available) and Jericho are domain-specific so agents that play these environments may perform well while conducting specific tasks like completing science experiments but lack generalized skills to apply them to other situations \cite{wang2022scienceworld,hausknecht2020interactive}. 
Most importantly, in order to generate a large number of games to train the RL agents to master skills in these environments, we will have to employ human annotators to manually design, generate, and deploy the game.
Therefore, the purpose of this work is to develop an efficient approach that generates a large amount of text-based games to train the RL agents to master the desired skills and excel at the target environments such as TWC, ScienceWorld, etc.

As developing a set of text-based games is a time-intensive manual process, we propose Self-supervised Text-bAsed RL learnING, "STARLING", an interactive environment that utilizes Large Language Models (LLM) and an integrated interactive fiction game engine (Inform7 \cite{inform7}) to easily produce games in any domain. 
We generate a set of $100$ text-based games using GPT3 \cite{NEURIPS2020_1457c0d6} based on the input game ideas (\textit{seed} list) that emphasize the need for the everyday skills such as \textit{boiling water,} \textit{cooking pasta,} \textit{etc}. in (pre-)training text-based RL agents. 
These games require agents to use a specific sequence of actions to achieve the goal and successfully complete the game. For example, while cooking pasta, an agent must first \textit{gather the ingredients}, \textit{fill pot} with water,  \textit{boil the water}, and \textit{put the pasta} in the pot. 
We then deploy the pre-trained RL agent on the target environments. 
This novel game-generation method can easily be used by others to create their own games and be adapted for future applications to build challenging RL agents in various domains. Figure \ref{fig:arch} shows the overview of the proposed approach for self-supervised text-based reinforcement learning using LLM.%

\section{Self-supervised Text-based RL}
Self-supervised RL involves bootstrapping RL agents with auxiliary tasks in an unsupervised or semi-supervised setting to accelerate learning and generalize in the target tasks. With the recent interest in LLMs, in this paper, we consider LLMs as an alternative option to pre-train an RL agent with minimal human supervision.  Unlike in the other text-based environments such as TextWorld \cite{cote2019textworld}, TextWorld Commonsense (TWC) \cite{murugesan2021text}, Jericho \cite{hausknecht2020interactive}, ScienceWorld \cite{wang2022scienceworld}, we utilize the skill generation capability of the large language models \cite{huang2022language} to automatically generate text-based games based on the input game ideas with minimal human supervision.%
Our proposed approach for self-supervised TBRL, STARLING is an interactive text-based environment with assistance from LLM and enables the text-based RL agent to hone their extra-curricular skills \footnote{In this work, we define the skills based on the auxiliary task such as "boil $<$object$>$", "fill $<$container$>$", "cook $<$object$>$", etc. We assume that LLM such as GPT3 has the necessary knowledge to generate text-based games based on these basic skills.}. In this paper, we assume a seed list of game ideas is already available as input to STARLING. These game ideas are chosen to exhibit specific skills either for creating a generalized agent or targeting domain-specific environments.  Optionally, the RL agent can generate a new set of game ideas during training, specific to the target domain to improve its performance.

\subsection{Constructing Pre-training Games}

\begin{figure}[t]
    \centering
    \includegraphics[width=0.88\linewidth]{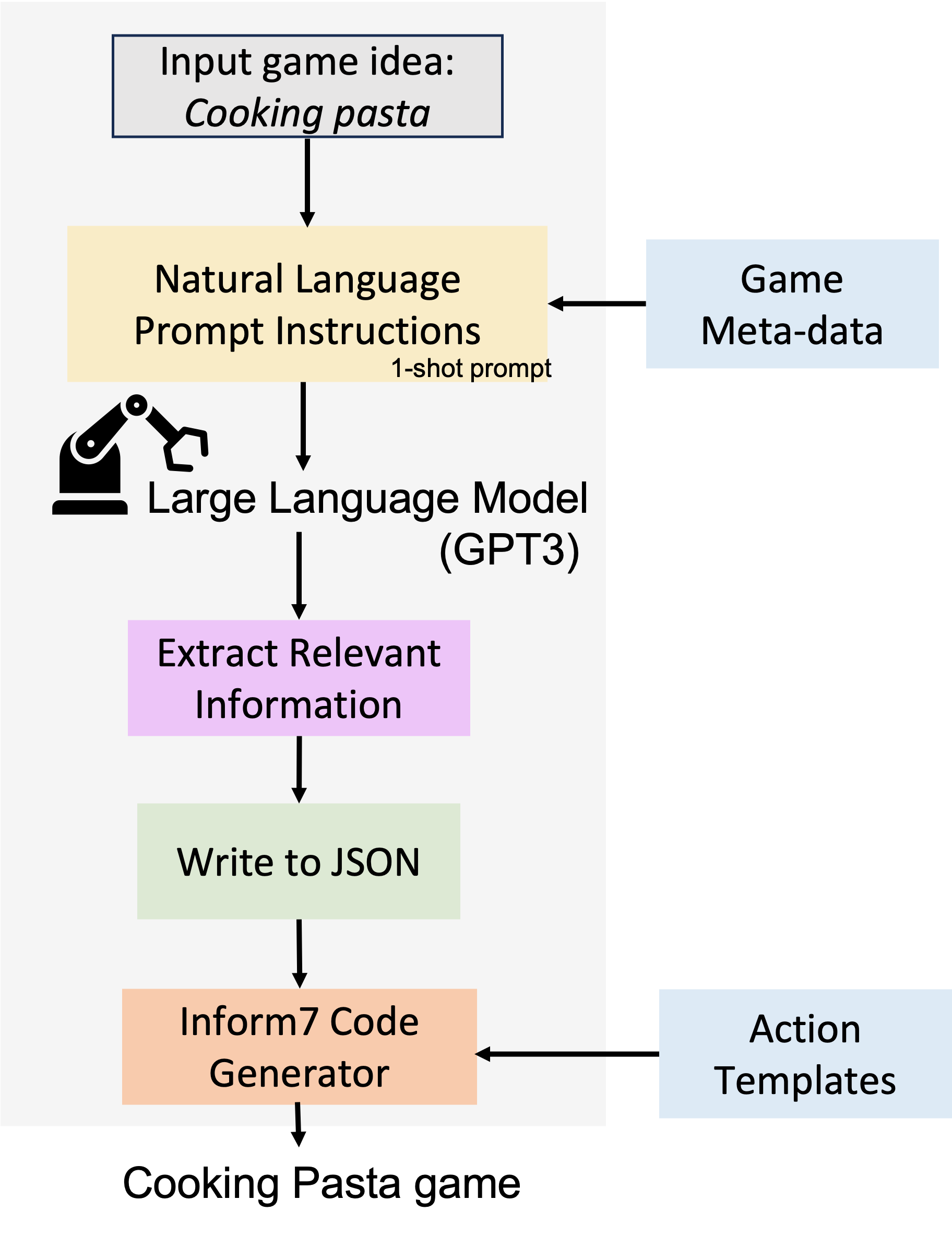}
    \caption{Workflow of the STARLING Game Generator using large language model (GPT3). }
    \label{fig:llm}
\end{figure}

\begin{figure*}[ht]
     \centering
     \includegraphics[width=0.985\linewidth]{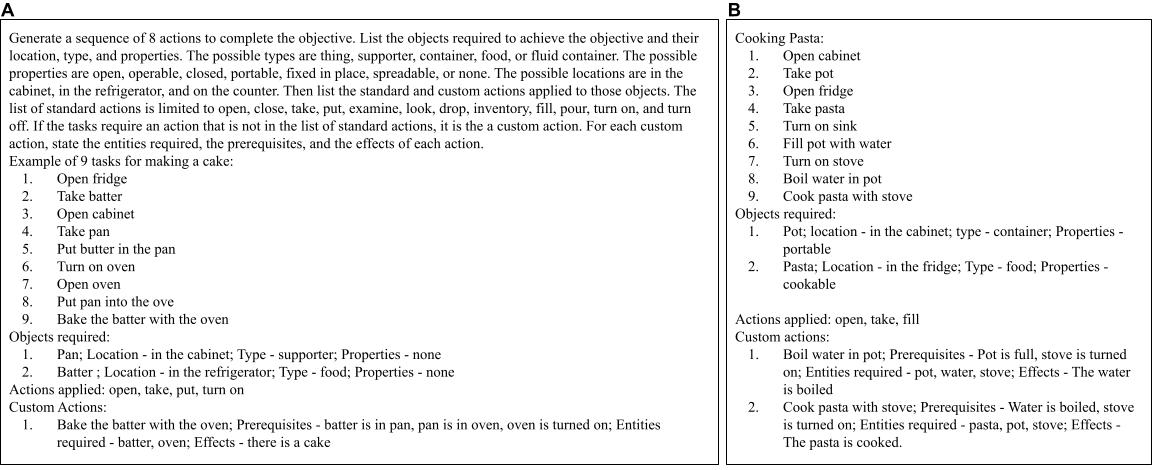}
     
    \caption{(A) GPT3 input prompt for cooking games with one action example. The actual prompt contains four action examples. (B) GPT3 output for cooking pasta game idea. GPT3 reliably outputs accurate and necessary game information very similar to the input.}
    \label{fig:example_output}
\end{figure*}
In this section, we briefly describe how we generate the pre-training games from the game ideas using LLM and Inform7.
Given the set of game ideas (seed) to LLM \cite{NEURIPS2020_1457c0d6}, we design a method that procedurally generates text-based games based on the interactive fiction game engine. In this paper, we use GPT3 as our LLM.
Inform7 is an interactive fiction programming language that allows users to create interactive fiction games using natural language instructions \cite{inform7}. Previous text-based environments such as TextWorld, Jericho, ScienceWorld, etc. use Inform7 (in the backend) to generate a handful of text-based games manually that require agents to explore the environment and take a sequence of actions to complete a goal such as \textit{cooking a pasta}. Based on our observation from these environments, we find that the game generation can be modularized into four parts: setup, object creation, custom action, and reward assignment:
\begin{enumerate}
    \item \textit{Setup} - defines basic properties about the game such as the room, entity types, any external libraries (Inform7), etc. 
    \item \textit{Object Creation} - creates in-game entities such as bread or jelly. Each entity is placed in its proper location like the refrigerator or cabinet and assigned properties such as portable, open, or closed.
    \item \textit{Custom actions} - defines actions not native to Inform7. Each custom action checks for the pre-conditions and then executes the action by initiating the relevant state changes, and returning the proper observations to the agent. We utilize predefined action templates to incorporate custom actions during the game generation.
    \item Rewards - assigns reward value for gathering the necessary entities and completing custom actions to achieve the goal. Once all the rewards are collected for each game, the game ends. 
\end{enumerate}

\begin{figure*}[ht]
     \centering
     \includegraphics[trim={35mm 0 35mm 0},clip,scale=0.235]{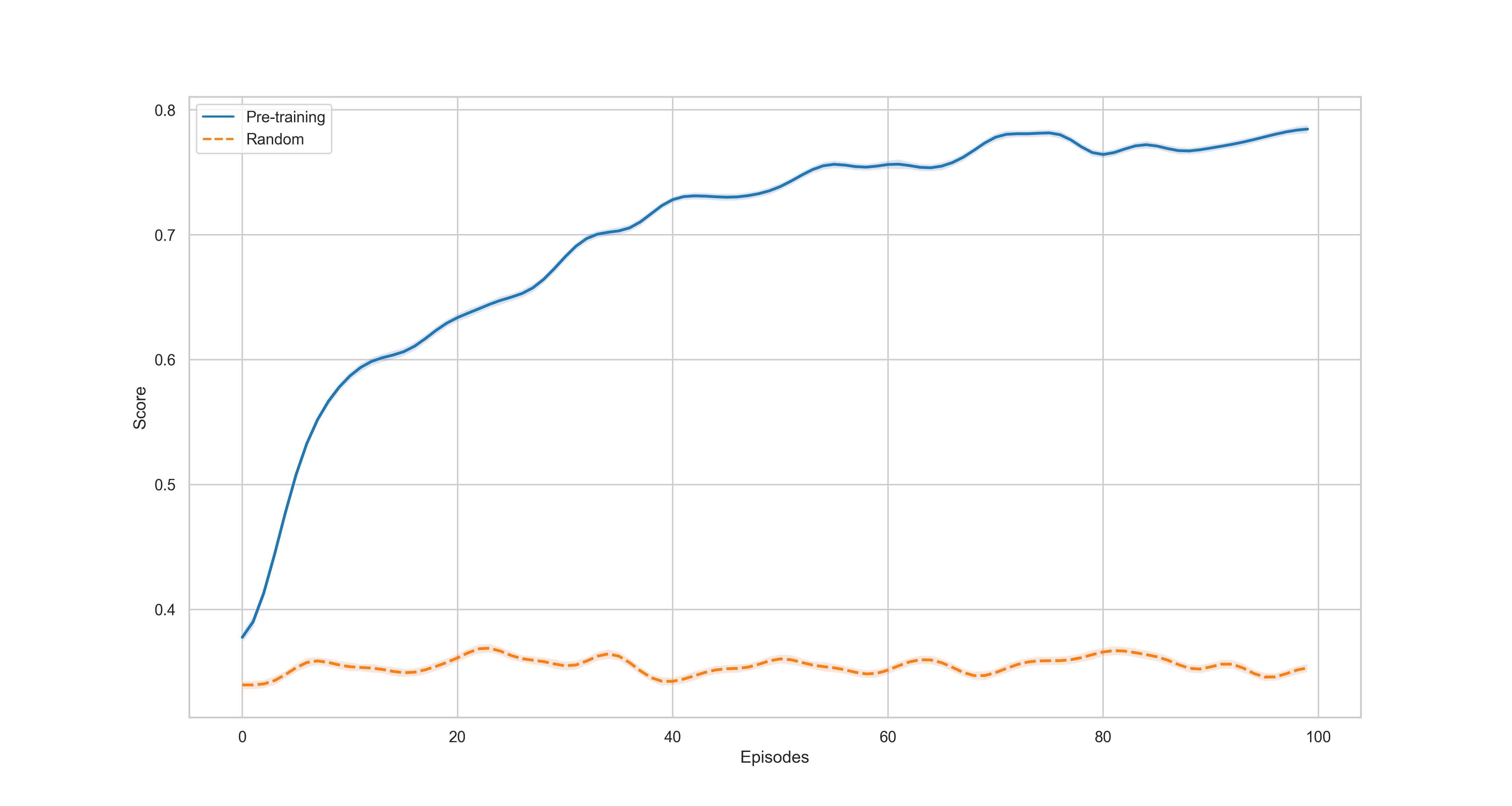}
     \includegraphics[trim={35mm 0 35mm 0},clip,scale=0.235]{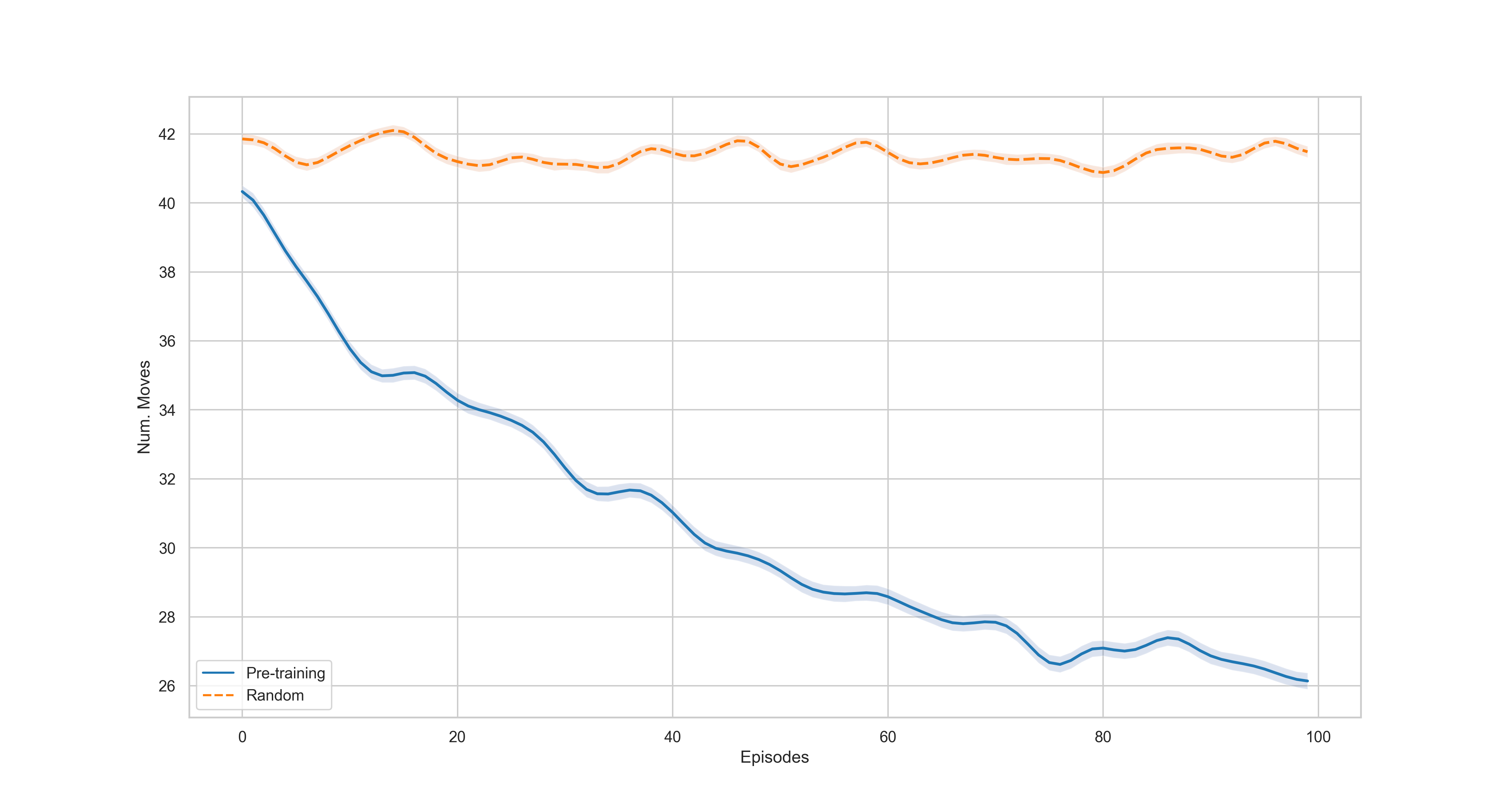}
    \caption{Training curves for pre-training step of STARLING depicting the normalized scores (left) and number of moves taken (right) of text-based reinforcement learning agents.}
    \label{fig:pre_training_curves}
\end{figure*}

Figure \ref{fig:llm} shows the overview of the game generation using STARLING. 
When we feed a game idea from the seed list, STARLING prepares a prompt using natural language instruction and example game metadata as shown in Figure \ref{fig:example_output}(a), with information about the setup, objects, custom actions, and rewards required for the game idea. We input this prompt to an LLM which generates the requested information as shown in Figure \ref{fig:example_output}(b). We initiate each prompt for a game idea with the necessary objects that the agent needs and agents must collect those objects and use them to cook, clean, build, or complete the high-level task. In order to be successful in these games, agents must understand the properties, location, and affordances of objects in addition to the specific sequence of actions needed to accomplish the task. 

 Next, we write the output from the GPT3 output into a JSON file as shown in Figure \ref{fig:json_file} (supplementary). The objects, actions, and tasks from the GPT3 output correspond to the entities, custom actions, and verbs sections of the JSON file respectively. At this stage, the user may update or change game information in the JSON file. If the user approves the game-related data in the JSON file, we write the Inform7 code based on the JSON file. We then convert this code into an Inform7 game for a given game idea using the Glulx \footnote{\url{https://en.wikipedia.org/wiki/Glulx}} interpreter for interactive fiction games. %

\subsection{Parsing LLM Response}

Since the response generated by LLM may not strictly follow the desired format, we follow additional steps to mitigate the irrelevant content in the response from LLM. 
First, we request a specific set of game-related data from LLM in a slot-filling style text generation to reduce the amount of long unstructured text generation \cite{rakotonirina2022can}.
Since LLM are good at instruction-following when few-shot examples (input-output pairs) of a similar problem are given as a part of the prompt input,
we add k-shot examples (k=3) to guide the LLM to generate a response. Figure \ref{fig:example_output}(a) shows one of the three examples given as a part of the input prompt.
Finally, during game compilation, the Glulx interpreter verifies whether the information extracted from the response adheres to the Inform7 programming language syntax.
In addition, the pipeline for game generation provides an option for users to review the generated JSON file. When the generated game files still contain irrelevant content, we repeat the text generation multiple times to get the desired response from LLM. 
 
\subsection{Game Insights}

Using the above approach repeatedly, we built a set of $100$ games with minimal human supervision for training and evaluating the text-based RL agents with skills. These games have on average $2$ skills and $4$ rewarded states per game.
These games have multiple sub-tasks which indicate that agents must utilize at least 2 skills (on average) for each game in the correct order. 

\begin{table}[h]
\centering
\resizebox{0.5\textwidth}{!}{%
\begin{tabular}{l|c|c}
\hline
Agents & Mean Normalized Score & Mean Moves Taken \\ \hline
Random      & 0.050 $\pm$ 0.01         & 50 $\pm$ 0.0    \\
Pre-training       & 0.72 $\pm$ 0.063         & 28.105 $\pm$ 1.876   \\
\hline
Human          & 1.000 $\pm$ 0.000         & 9.640  $\pm$ 5.620   \\  \hline
\end{tabular}
}
\caption{
Performance of random and pre-trained agents on a set of 25 unseen pre-training games after training on 75 pre-training games over 100 episodes. Mean Norm. Score (higher is better, normalized with maximum score achievable per game) and Mean Moves Taken (to achieve the goal, lower is better).
}
\label{tab:pre_results}
\end{table}
Agents must take approximately $7$ actions on average to complete each game, though some of these actions do not necessarily have to be completed in order (e.g. the agent can "turn on the stove" before "fill the pot with water" and vice versa). Games in TWC only require agents to gather objects and take actions to place them in their commonsense locations. These actions can often be completed in any order, whereas generated games, such as \textit{cooking pasta}, require agents to gather objects and use other related skills in a specific sequence to achieve the final goal.

\section{Experiments}
In this section, we report the experimental results of the proposed approach: STARLING. We pre-train the RL agent on the generated $100$ games (train split). We evaluate the pre-trained agent (STARLING) on three benchmark environments for text-based games: TextWorld Commonsense (TWC) with easy, medium, and hard difficulty levels \cite{murugesan2021text} 2) ScienceWorld with 4 tasks and variations \cite{wang2022scienceworld}. 3) Zork1 from Jericho.

\begin{figure*}[ht]
     \centering
     \includegraphics[trim={30mm 0 30mm 0},clip,width=0.325\textwidth]{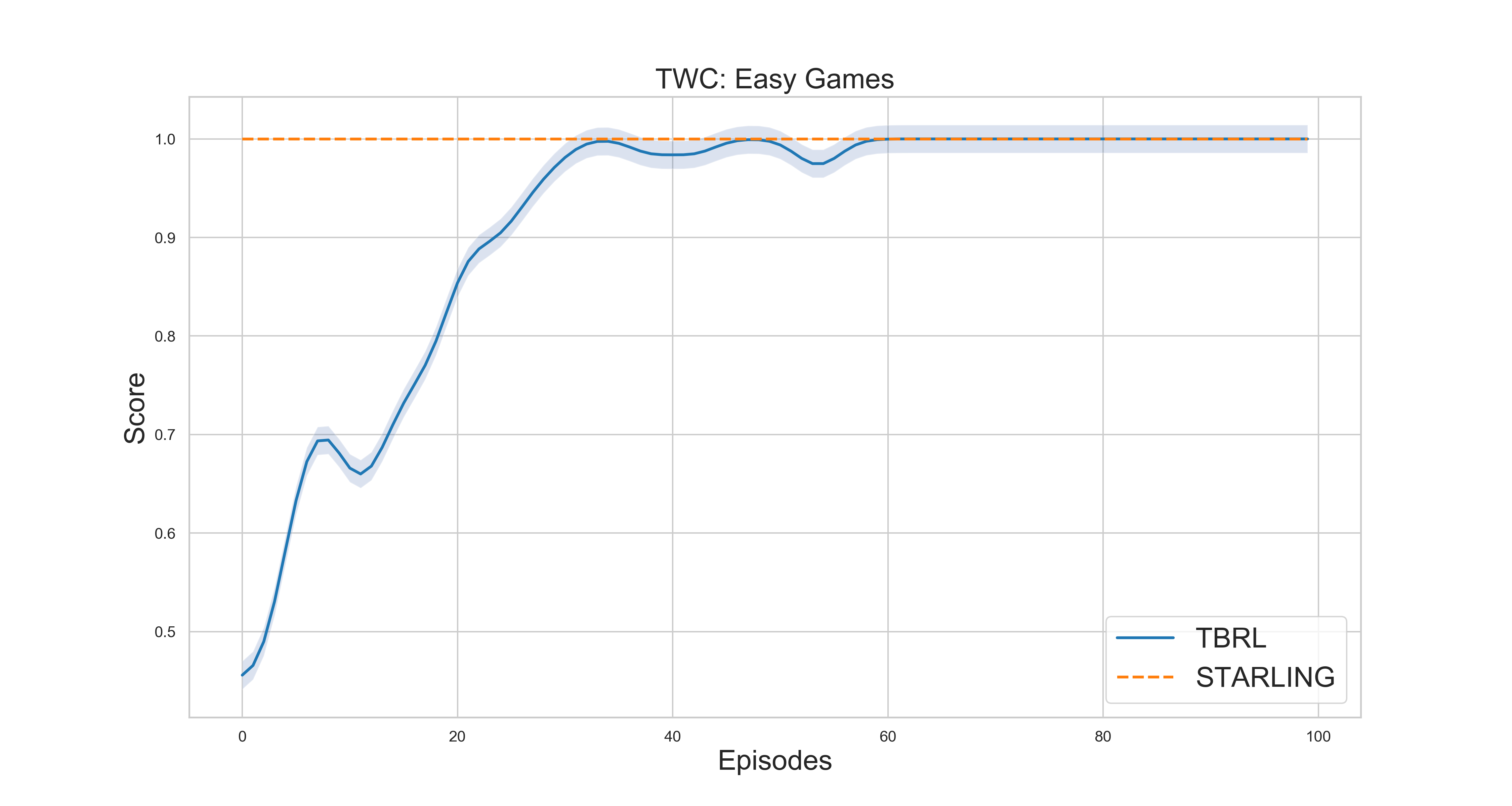}
     \includegraphics[trim={30mm 0 30mm 0},clip,width=0.325\textwidth]{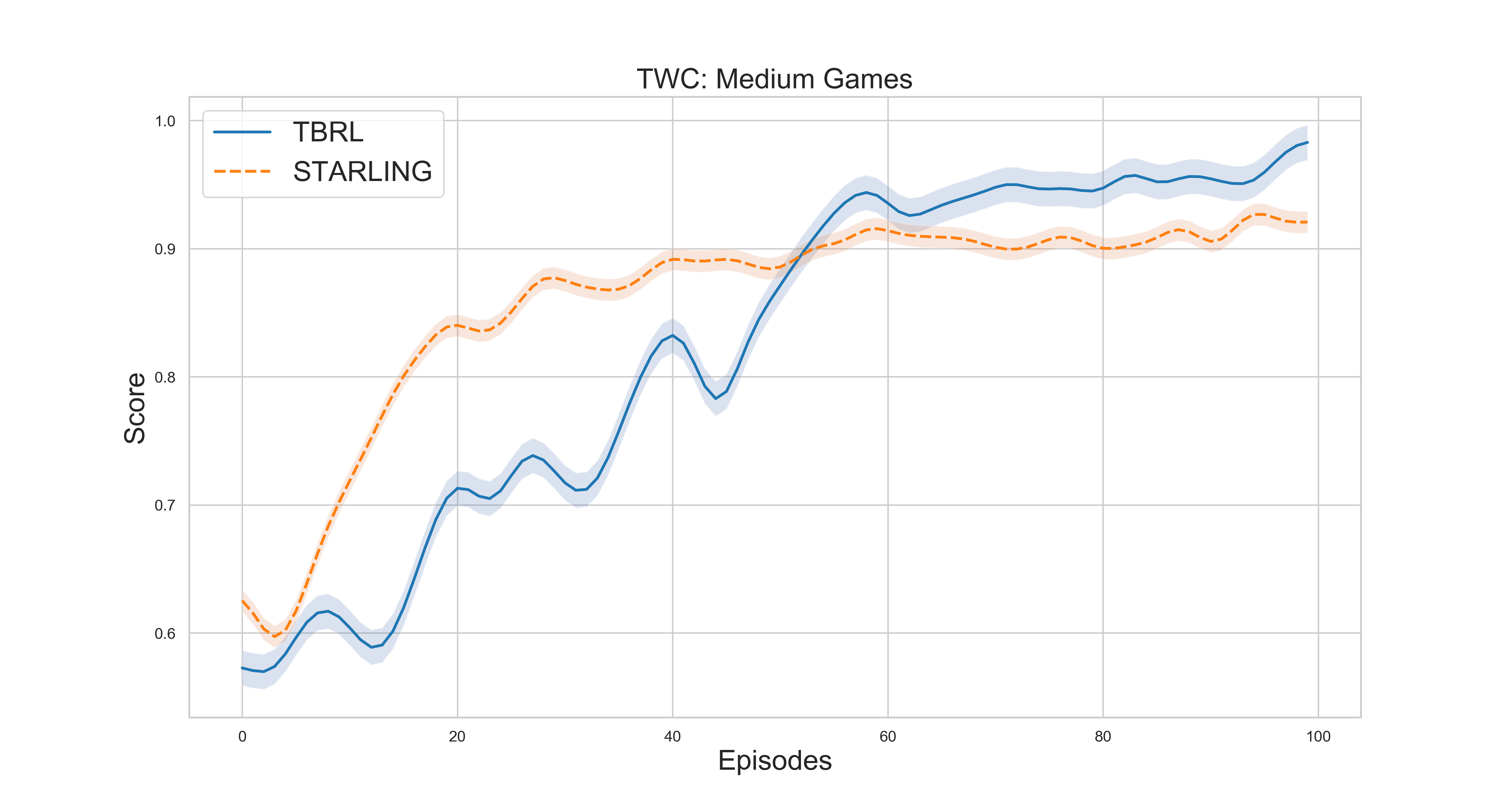}
     \includegraphics[trim={30mm 0 30mm 0},clip,width=0.325\textwidth]{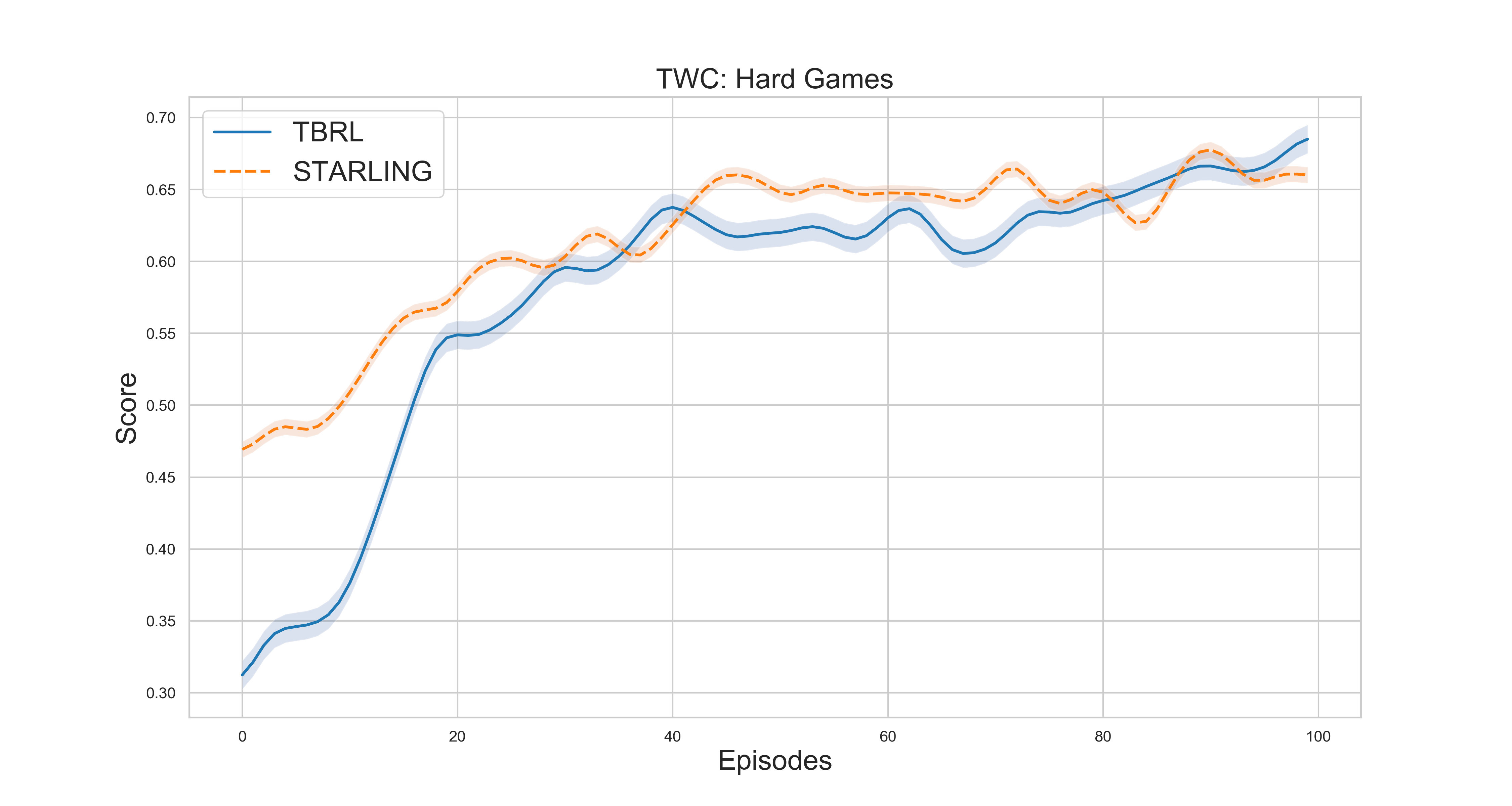}\\
     \includegraphics[trim={30mm 0 30mm 0},clip,width=0.325\textwidth]{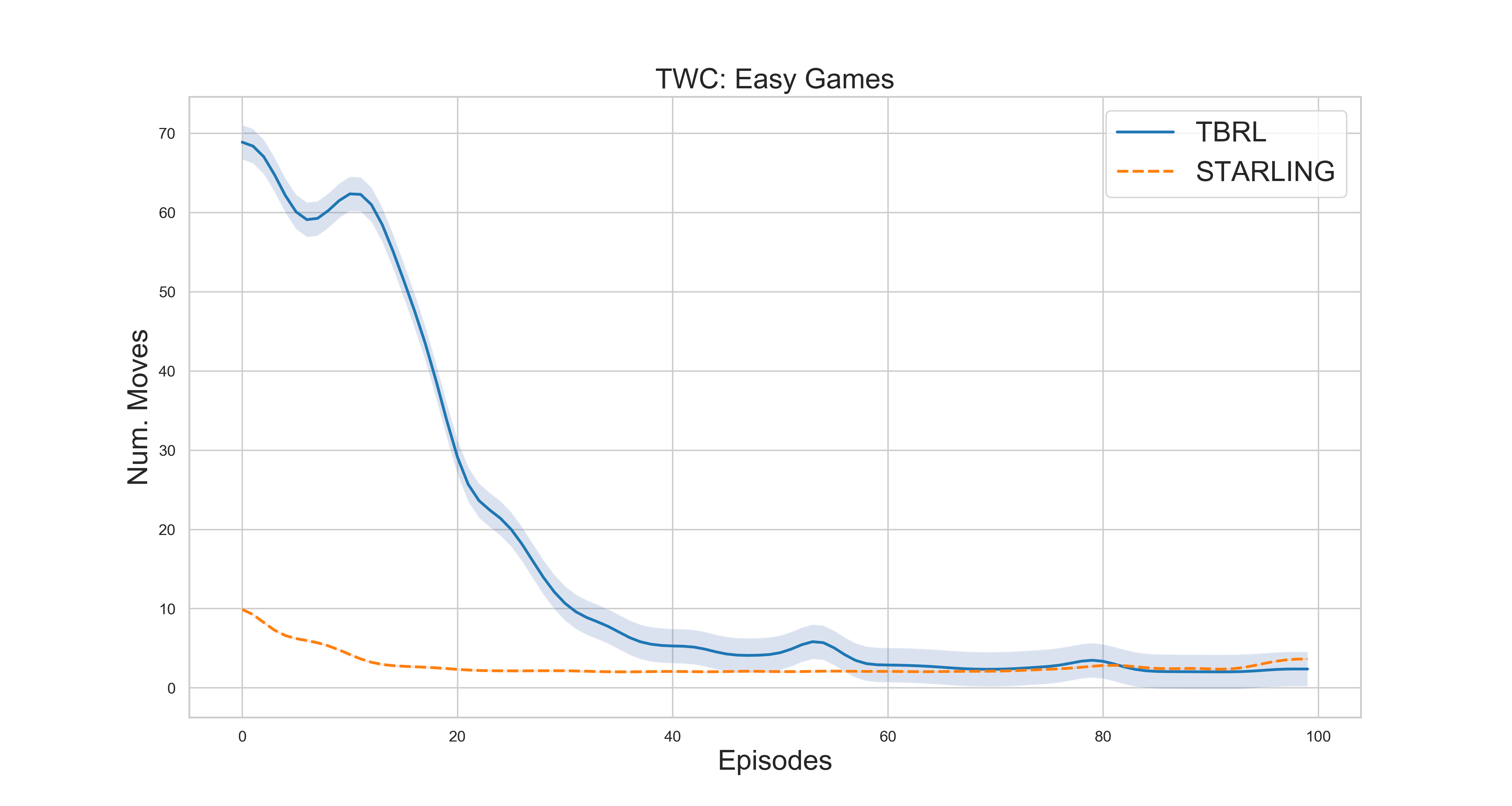}
     \includegraphics[trim={30mm 0 30mm 0},clip,width=0.325\textwidth]{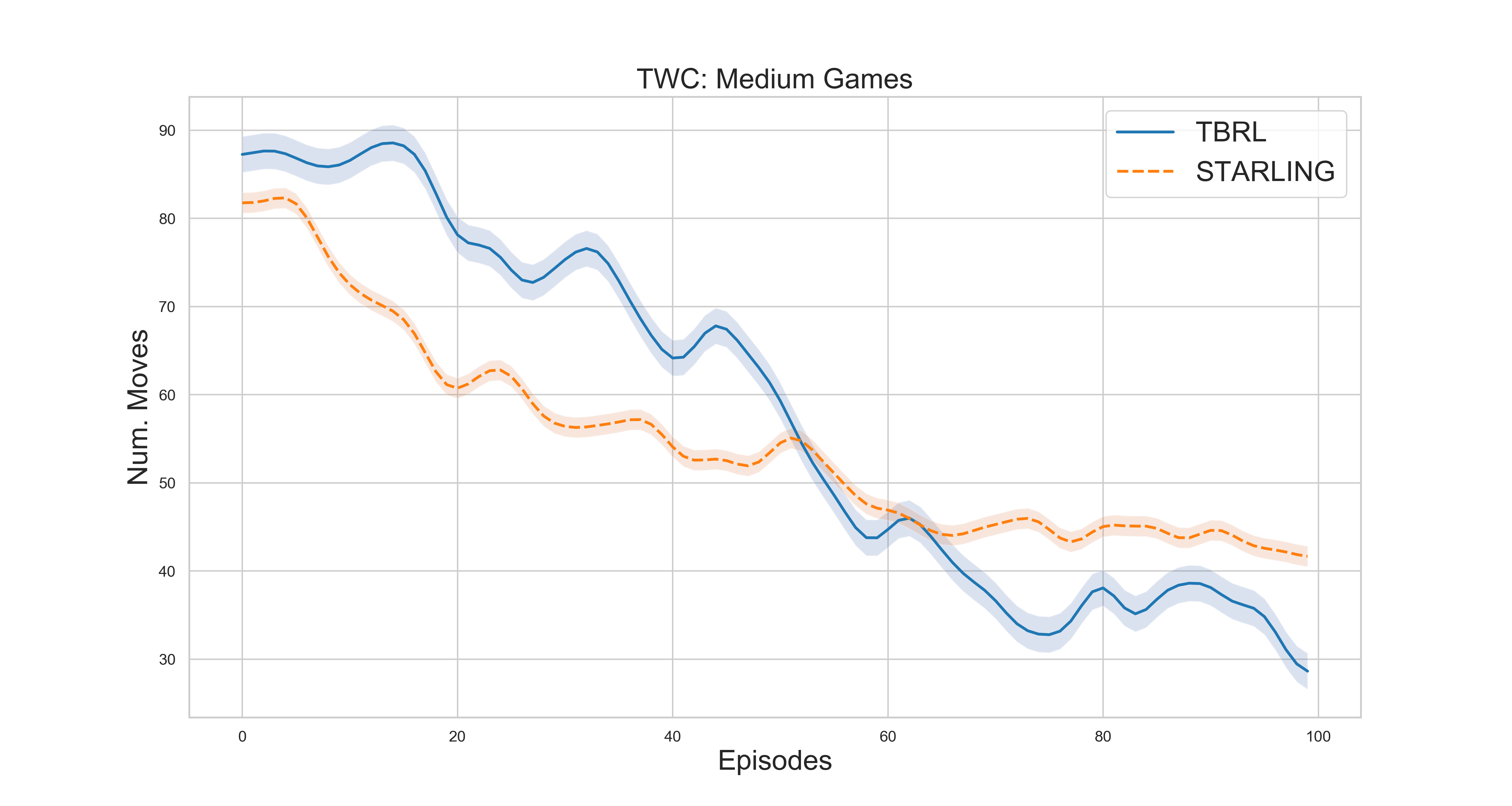}
     \includegraphics[trim={30mm 0 30mm 0},clip,width=0.325\textwidth]{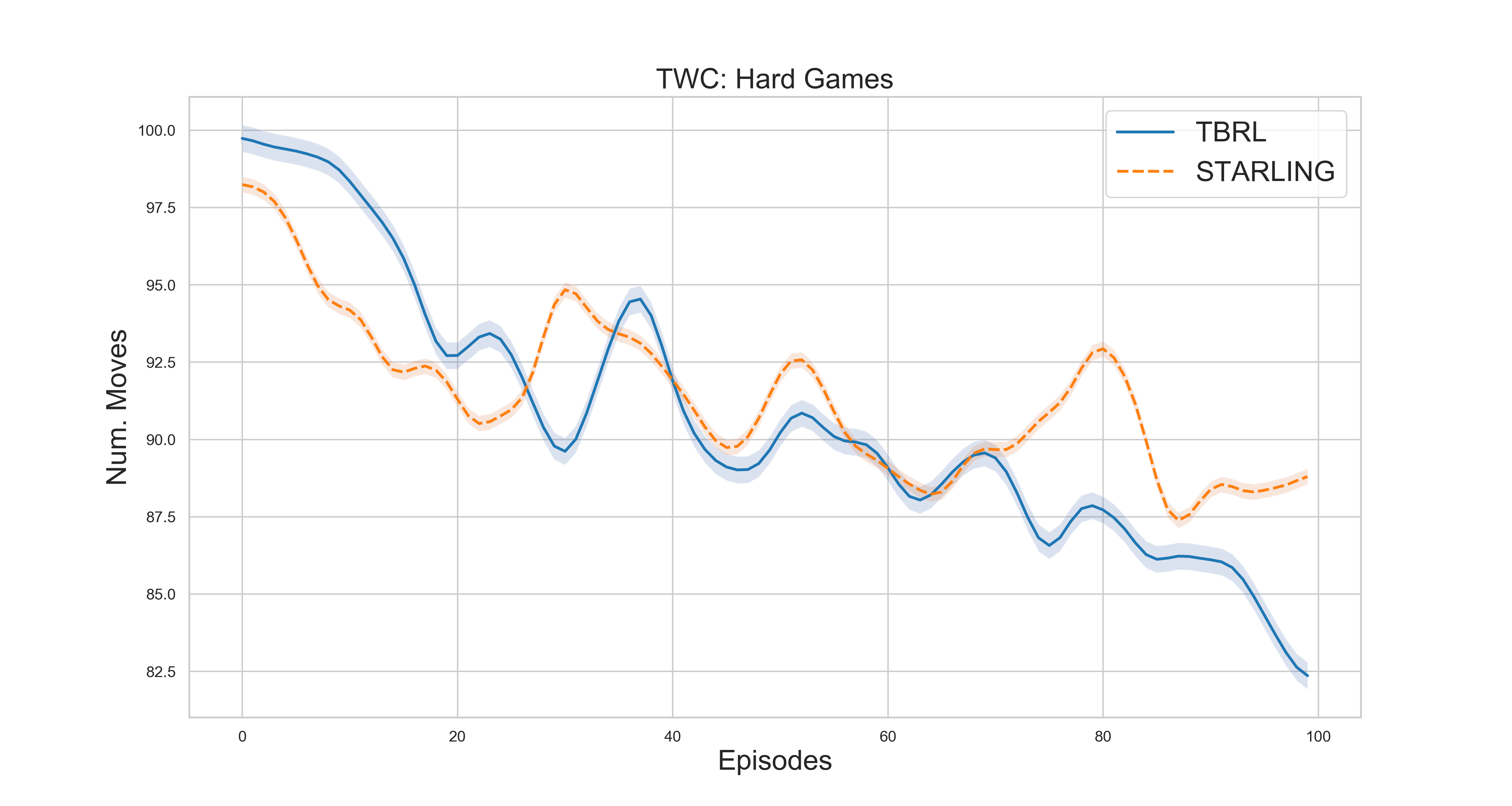}
    \caption{Training curves for TWC easy (left), medium (middle), and hard (right) games depicting the normalized scores (top) and number of moves (bottom) of both vanilla TBRL and STARLING agents.}
    \label{fig:twc_training_curves}
\end{figure*}
\begin{table*}[ht]
\centering
\resizebox{0.55\textwidth}{!}{%
\begin{tabular}{l|lll}
\hline
Agents &
  \multicolumn{3}{c}{Textworld Commonsense} \\ \hline
 &
  \multicolumn{1}{l|}{Easy} &
  \multicolumn{1}{l|}{Medium} &
  Hard  \\ \hline
\textit{Vanilla TBRL} &
  \multicolumn{1}{l|}{0.99 $\pm$ 0.0} &
  \multicolumn{1}{l|}{0.81 $\pm$ 0.06} &
  0.57 $\pm$ 0.03  \\ \hline
\textit{STARLING} &
  \multicolumn{1}{l|}{\textbf{1.0 $\pm$ 0.0}} &
  \multicolumn{1}{l|}{\textbf{0.85 $\pm$ 0.04}} &
  \textbf{0.64 $\pm$ 0.05} \\ \hline
\end{tabular}%
}
\caption{Performance comparison of vanilla TBRL (without pre-training) and STARLING on the test set with the three difficulty levels of Textworld Commonsense (TWC). All the scores and moves are averaged over $3$ runs.}
\label{tab:final_result_twc}
\end{table*}

\subsection{Text-based RL Agent}
In this section, we briefly describe the text-based agent used for all the experiments. Based on the recent observation that using LLM to learn the underlying representation of text in the environment does not necessarily improve the performance  \citep{wang2022scienceworld}, we follow previous works \cite{he2016deep,murugesan2021text,ammanabrolugraph,yao2020keep,atzeni2022case} and use GRU-based \textit{Vanilla TBRL} agent to evaluate our proposed approach (Figure \ref{fig:tbrl} supplementary). We use individual GRUs \cite{cho2014learning} for the information from the text-based environment such as observed text, the content of the inventory, the description of the room where the agent currently is located, and a valid list of actions.
We learn the state representation by concatenating the individual representations from their GRUs \cite{cho2014learning}. We compute the action probability from both the state and action representations.
We use Advantage Action-critic (A2C) to train the network \cite{mnih2016asynchronous}. In order to limit the impact of architecture and text-based RL algorithms, we consider these standard architectures for representation learning and RL algorithms that have proven to work well in text-based environments.  We plan to study the impacts of different RL algorithms and diverse sets of representation learning approaches in our future work. 
All the results reported in this paper are averaged over $3$ runs.

\subsection{Pre-training Text-based RL Agent}

\begin{figure*}[ht]
     \includegraphics[width=0.552\textwidth]{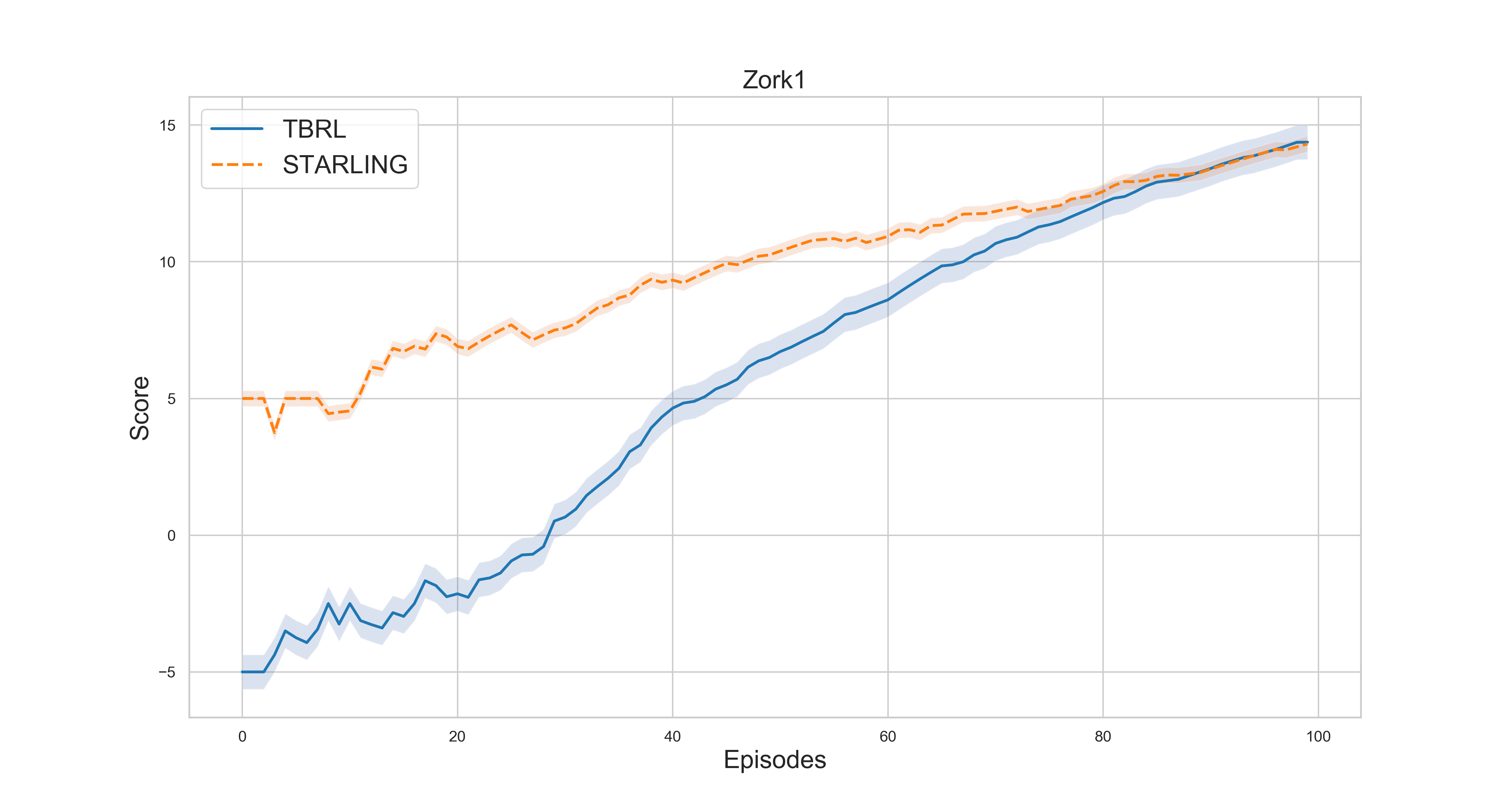}
     \includegraphics[width=0.454\textwidth]{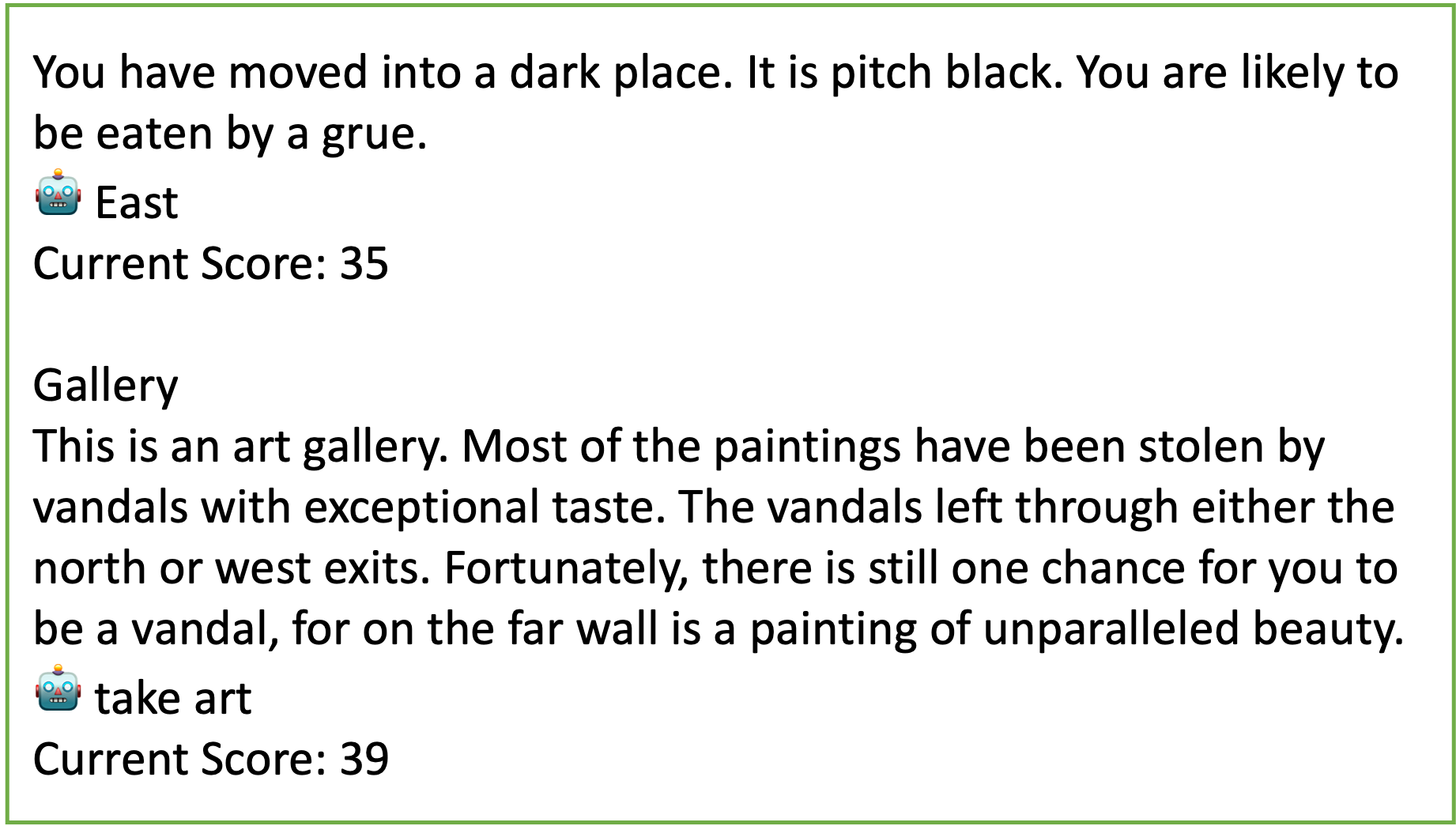}
    \caption{(left) Performance of TBRL agents on Zork1. (right) Sample trajectory from STARLING showing the bonus points of (+4 scores) for collecting the painting in the art gallery within the first few episodes.}
    \label{fig:zork_results}
\end{figure*}

We generate $100$ games that demonstrate basic skills in these environments such as \textit{cooking pasta, painting the living room, boiling water, lighting a candle,} etc. We split these $100$ games into $75$ games for training and $25$ for evaluation.
 We trained the vanilla TBRL agent on these $75$ pre-training games over $100$ episodes ($50$ max. steps per episode) and evaluated it on $25$ held-out games. 
We compare the performance of the vanilla TBRL agent (pre-training) against both the random agent (picks random action at each step) and the Human performance. We use the mean normalized score and mean moves/steps taken by the agent for comparison. We collect the Human performance results based on the $48$ participants (Section \ref{sec:human}). 
Figure \ref{fig:pre_training_curves} shows the training performance and Table \ref{tab:pre_results} shows the evaluation results. We can see that human participants (high-school students) solved these games with a perfect normalized score of $1.0$ indicating that these games are easy to solve. In order to successfully finish a game, an agent needs to take certain actions in a particular order. The order of actions taken decides the future states of the entities involved in the game.

\subsection{Self-supervised Training of Text-based RL Agent}
Next, we deploy the TBRL agent pre-trained on $75$ games generated by LLM on different environments. We call the pre-trained TBRL agent STARLING. We expect that STARLING will outperform the vanilla TBRL agent by utilizing the skills learned using LLM and boosting the performance and generalization capabilities to reach the goal of the target environments: ScienceWorld, TWC, and Zork1.

\subsubsection{TextWorld Commonsense Environment}
TextWorld Commonsense environment evaluates the agent on commonsense reasoning about everyday objects such as toothbrush, dirty towel, etc. The environment, based on Textworld engine \cite{cote2019textworld}, includes three difficulty levels: easy, medium, and hard depending on the number of objects to find and the number of rooms to explore. Each difficulty level includes $5$ training games and $5$ evaluation games similar to the distribution of the training games\footnote{In addition to the $5$ evaluation games, TWC includes $5$ test games from out-of-distribution. Since these games require external knowledge such as ConceptNet \cite{speer2017conceptnet}, ATOMIC \cite{hwang2021comet} etc., we exclude them from our experiments.} for a total of $30$ games with a batch size of $1$ for this experiment. We train the STARLING agent on these $15$ games for 100 episodes with a maximum of $50$ steps.

Figure \ref{fig:twc_training_curves} shows the training curves of the three difficulty levels in the TWC environment. We can see that the STARLING agent gets a boost in performance both in the scores achieved and the moves taken compared to the vanilla TBRL. This shows that the pre-training step using LLM in STARLING leverages the basic skills mastered using the $75$ generated games.  Table \ref{tab:final_result_twc} confirms our hypothesis that the pre-training step in STARLING improves the overall performance across different difficulty levels.

\begin{figure*}[h!]
     \centering
     \includegraphics[trim={30mm 0 30mm 0},clip,width=0.45\linewidth]{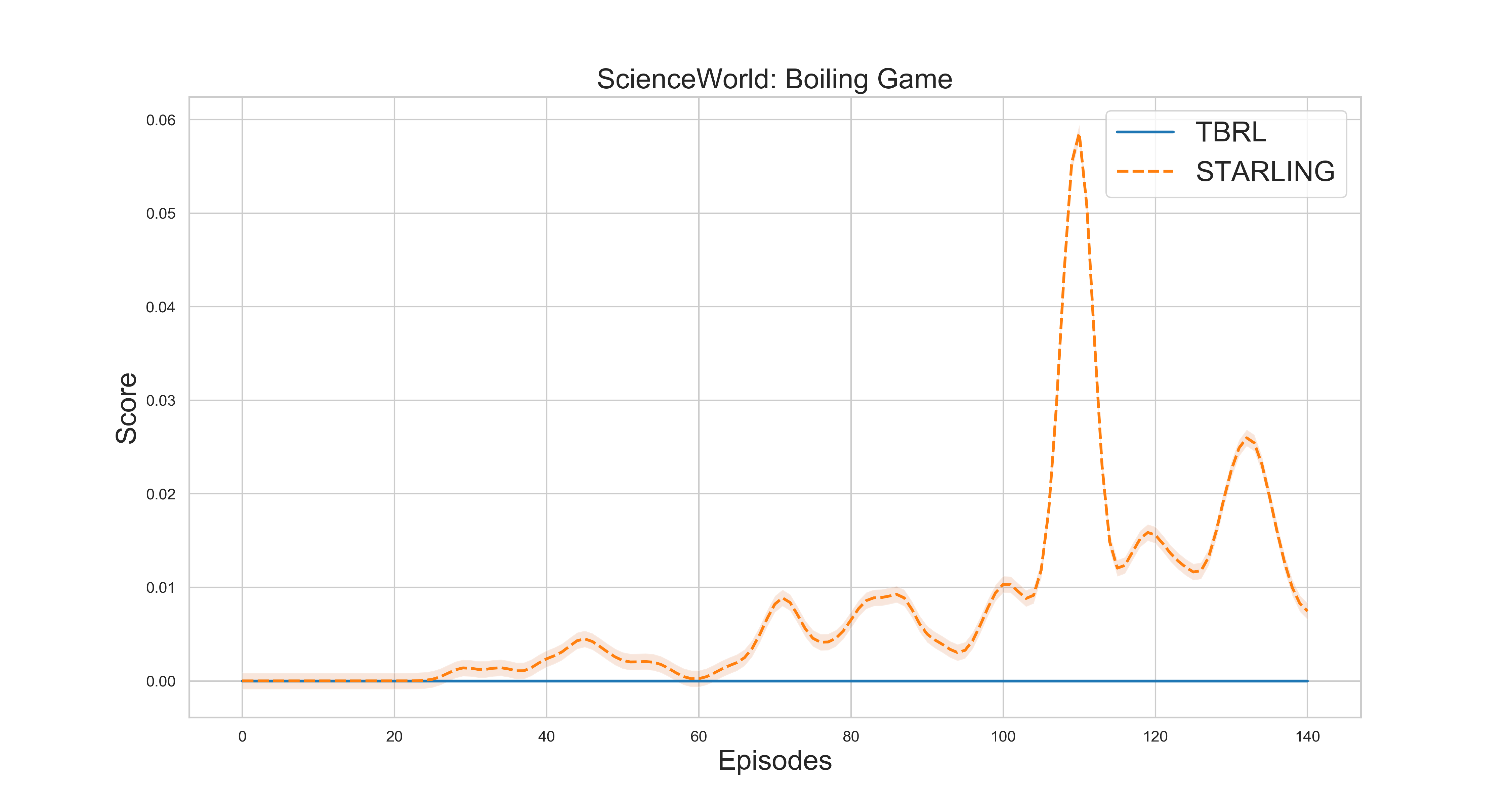}
     \includegraphics[trim={30mm 0 30mm 0},clip,width=0.45\linewidth]{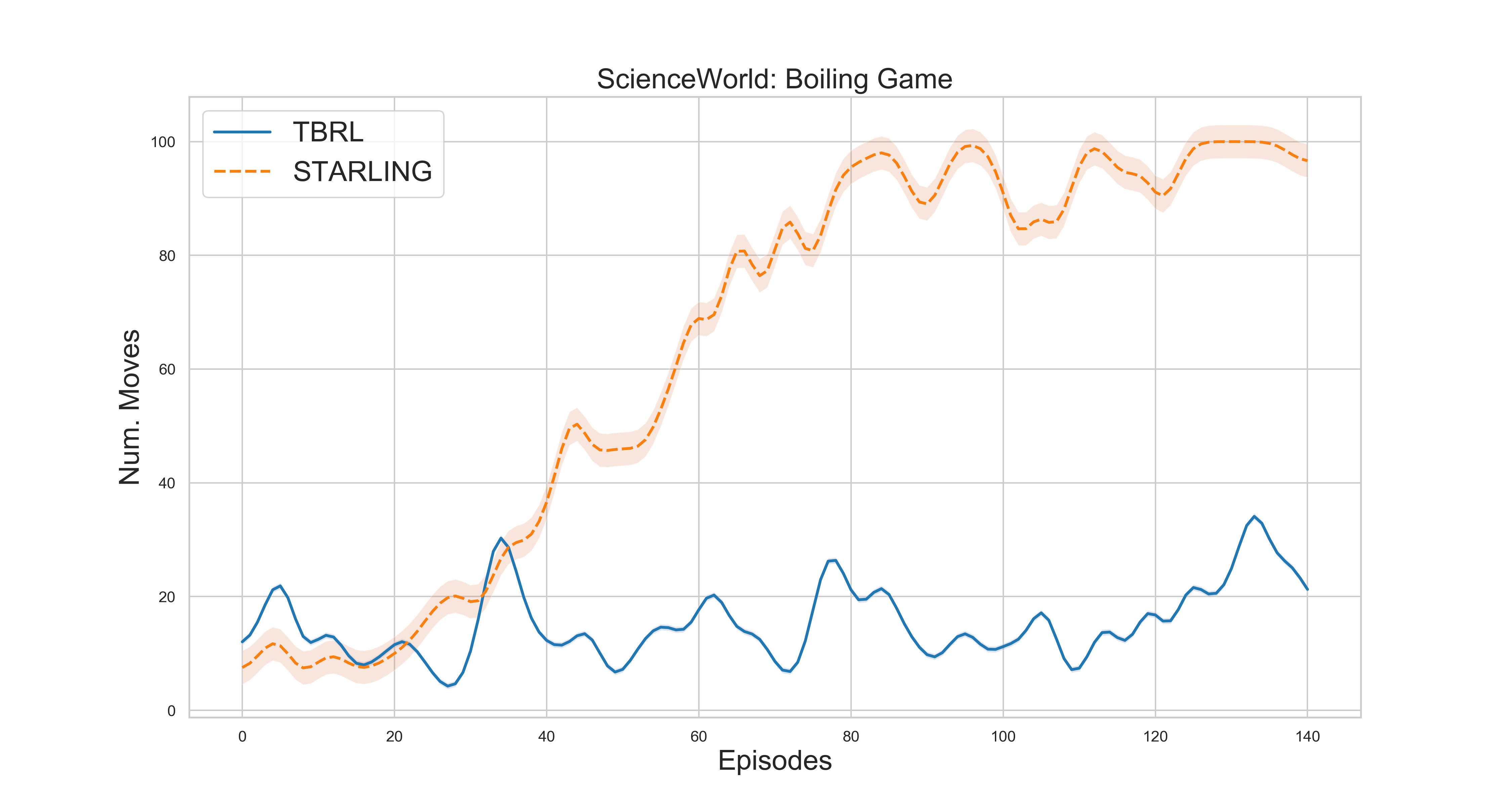} \\
     \includegraphics[trim={30mm 0 30mm 0},clip,width=0.45\linewidth]{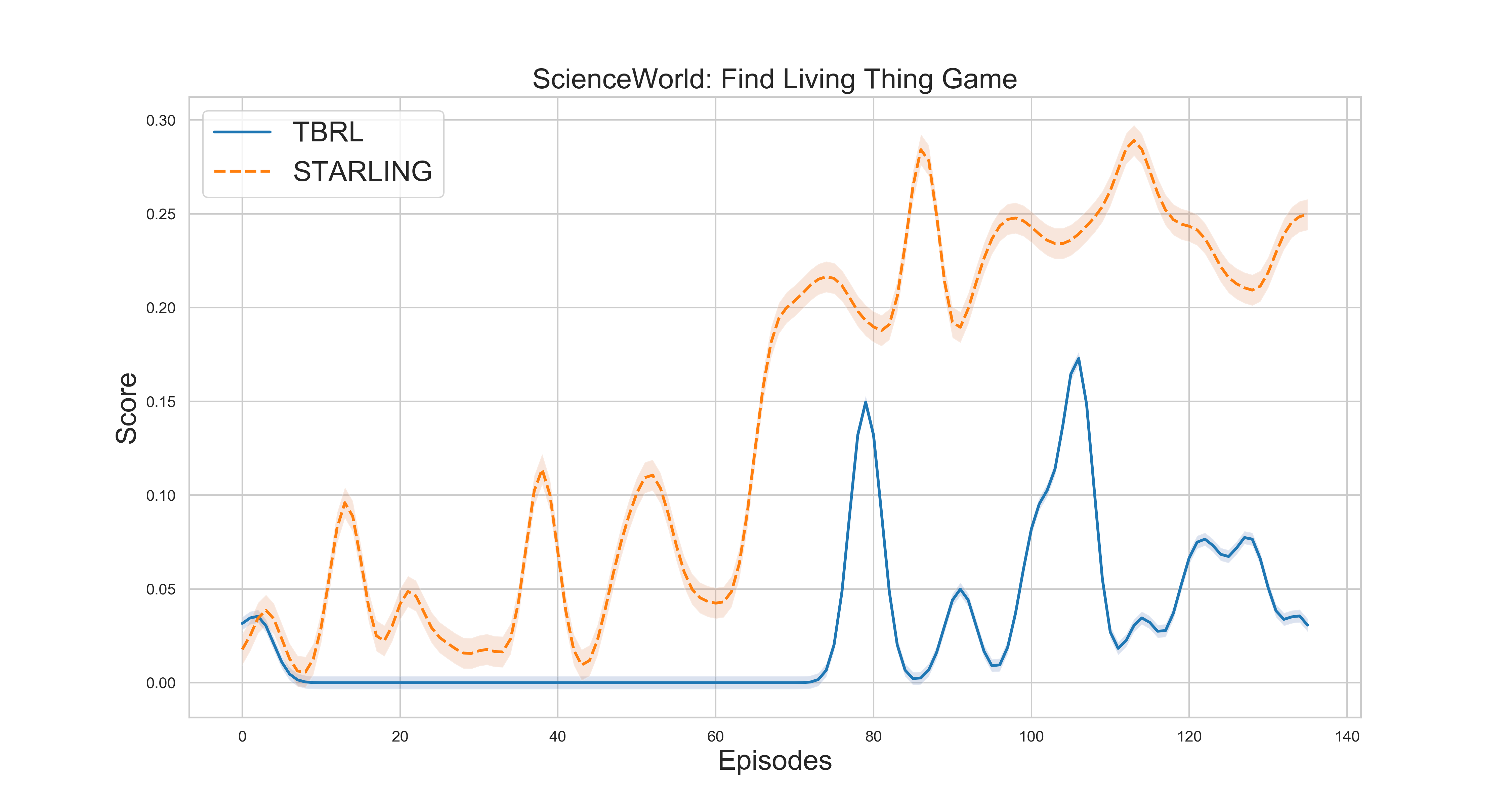}
     \includegraphics[trim={30mm 0 30mm 0},clip,width=0.45\linewidth]{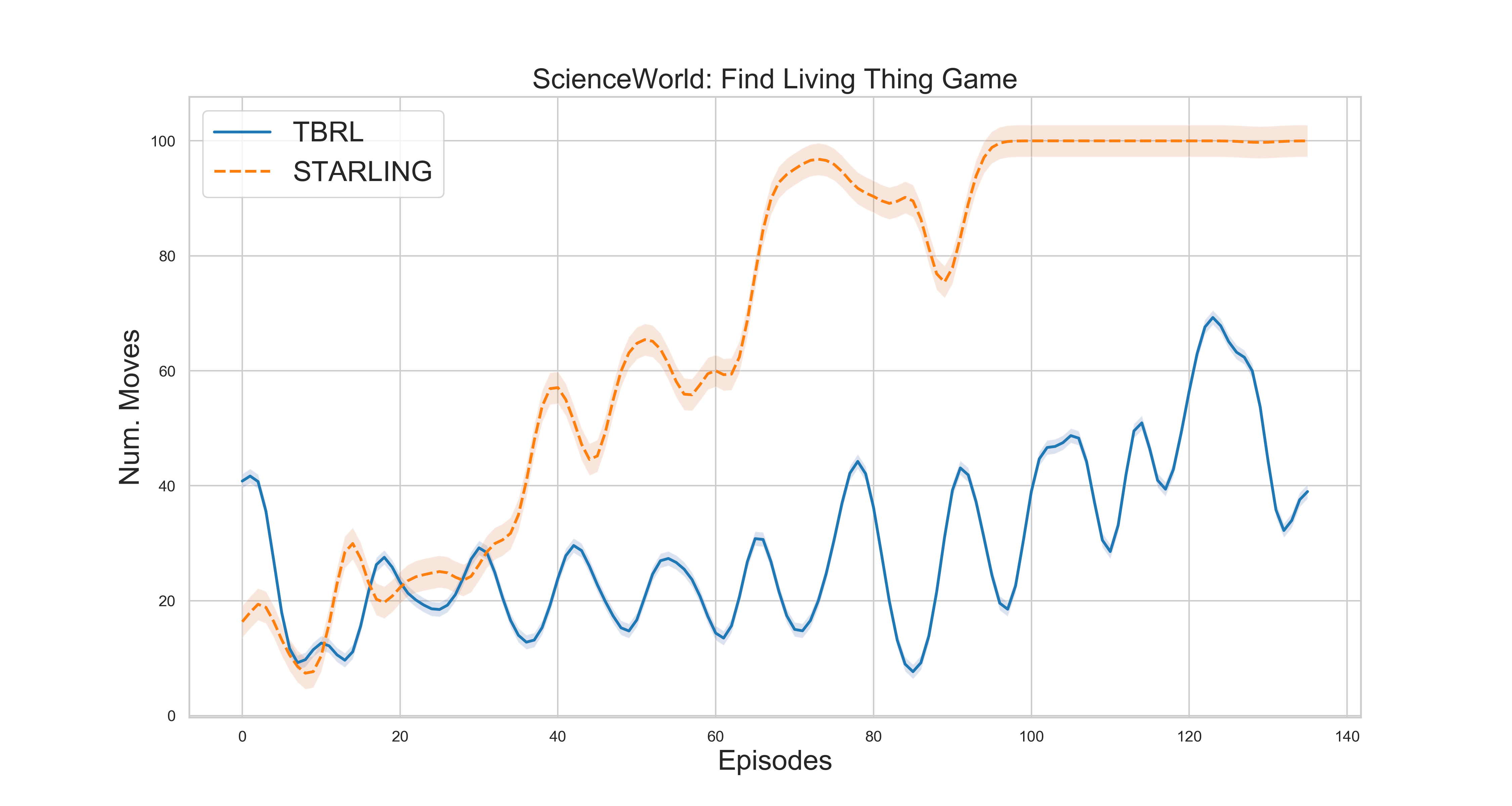}\\
     \includegraphics[trim={30mm 0 30mm 0},clip,width=0.45\linewidth]{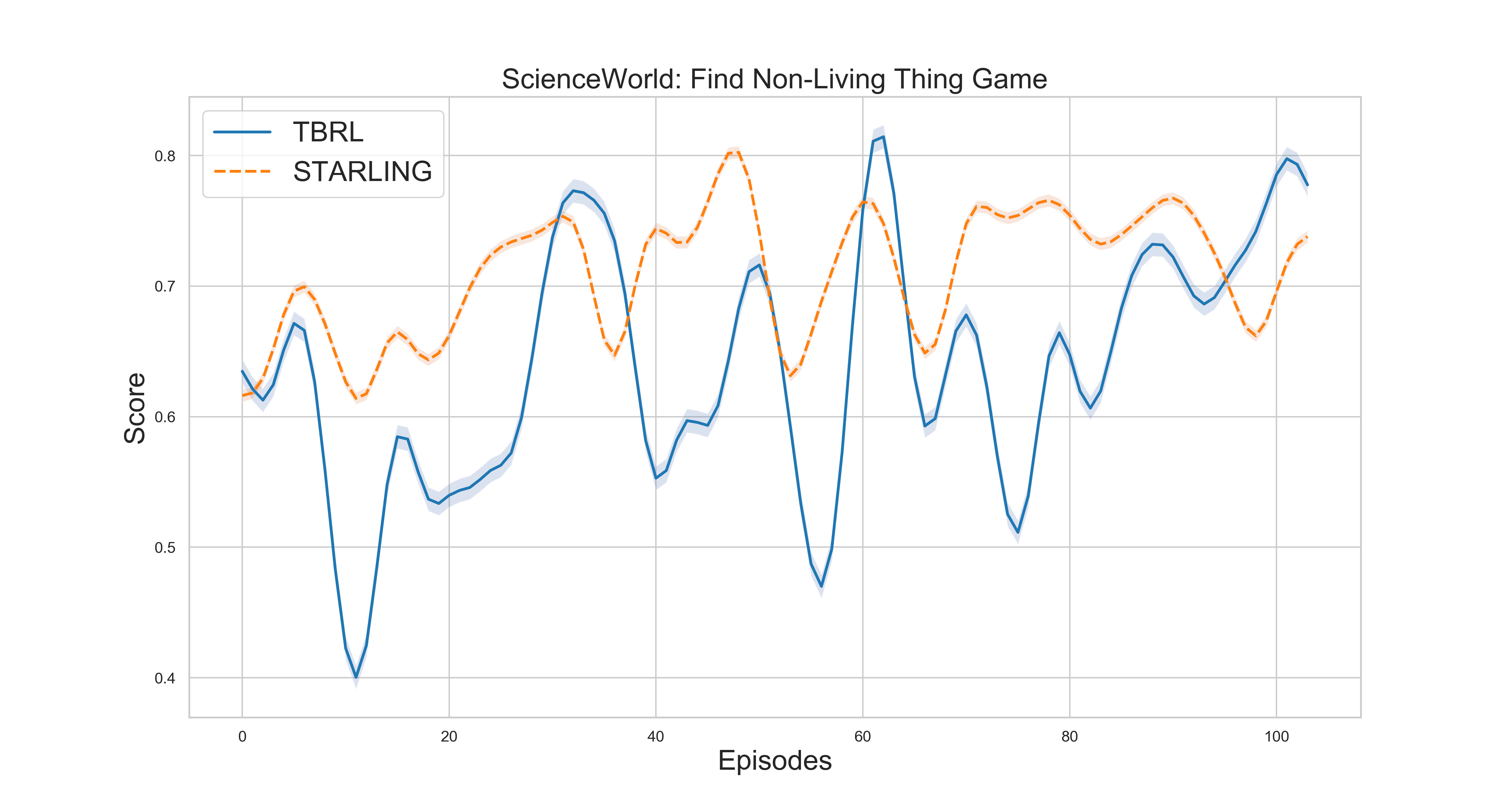}
     \includegraphics[trim={30mm 0 30mm 0},clip,width=0.45\linewidth]{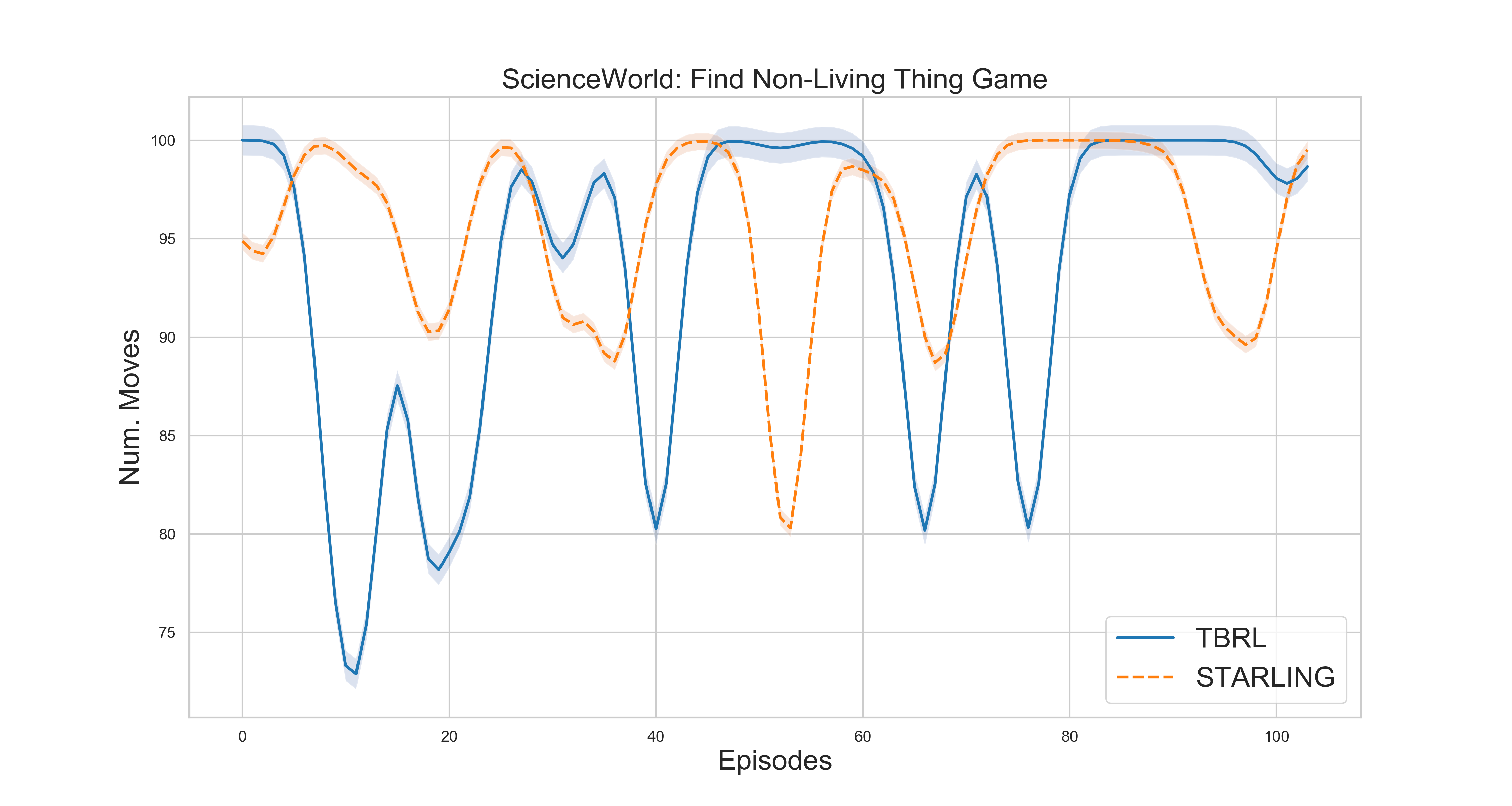}\\
     \includegraphics[trim={30mm 0 30mm 0},clip,width=0.45\linewidth]{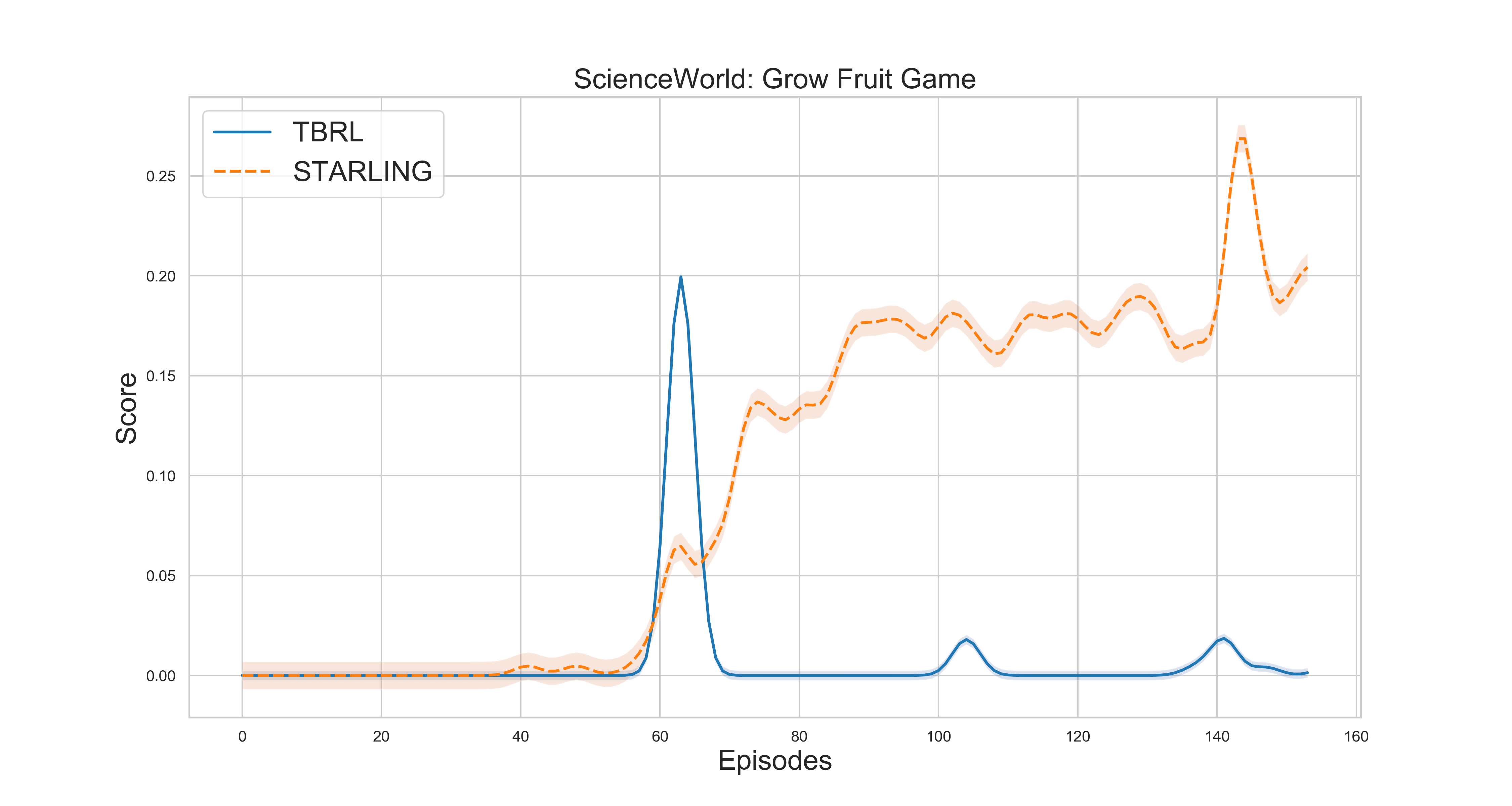}
     \includegraphics[trim={30mm 0 30mm 0},clip,width=0.45\linewidth]{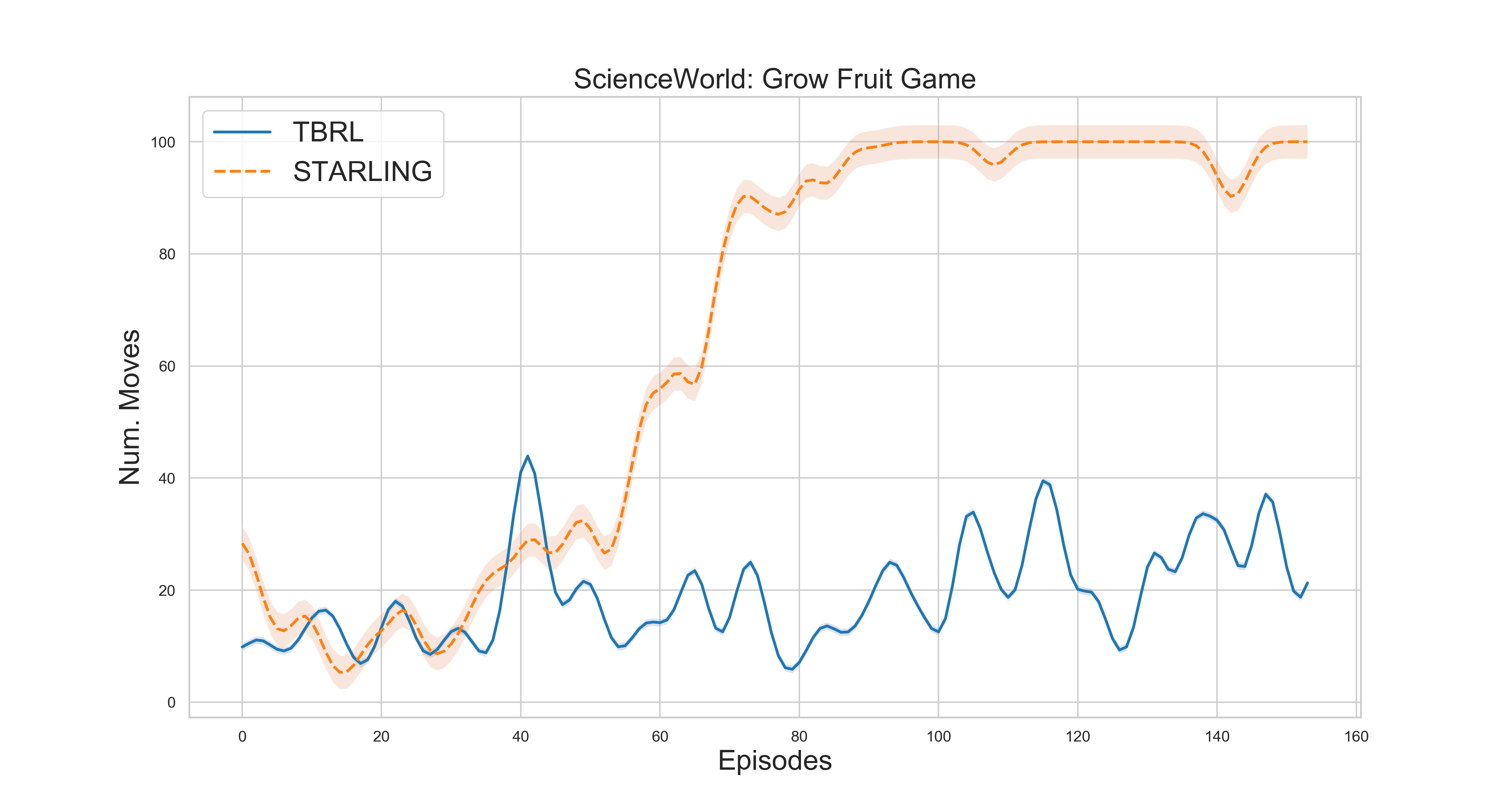}
    \caption{Training curves for ScienceWorld - Boil Substance (Matter), Find a living thing (Classification), Find a non-living thing (Classification), and Grow a fruit (Biology) games depicting the scores (left) and number of moves (right) of text-based reinforcement learning agents. }
    \label{fig:sw_training_curves}
\end{figure*}

\begin{table*}[!t]
\centering
\scriptsize
\vspace{8mm}
\begin{tabular}{llccccccccc}
\textbf{Theme} & \textbf{Task} &
Random-Valid\textsuperscript{*} & DRRN\textsuperscript{*} & KG-A2C & CALM & BC\textsuperscript{*} & TDT\textsuperscript{*} & STARLING\textsuperscript{*} \\
\toprule
     \multicolumn{2}{c}{\textbf{Model Types}}  &                   --                 & GRU      &  GRU   & GPT2  &   T5 (base)  &  
     T5 (base) & GRU  \\
    \multicolumn{2}{c}{\textbf{Param. Count $\times10^6$}}                                     &      &  {1.5}   &    5.5     &  131  &   11,000  & 11,000  & 0.2  \\
\toprule
\rowcolor[HTML]{f1f1f1}
Matter         & Changes of State (Boiling)                  & 0.00 & 0.03 & 0.00 & 0.00 & 0.00 & 0.00 & \textbf{0.04}\\ %
\rowcolor[HTML]{f1f1f1} 
Matter         & Changes of State (Melting)                  & 0.00 & 0.04 & 0.00 & 0.00 & 0.00 & 0.01  & \textbf{0.05}\\ %
\rowcolor[HTML]{f1f1f1} 
Matter         & Changes of State (Freezing)                 & 0.00 & 0.01 & 0.04 & 0.00 & 0.01 & 0.00  & \textbf{0.10}\\ %
\rowcolor[HTML]{f1f1f1} 
Matter         & Changes of State (Any)                      & 0.00 & \textbf{0.03} & 0.00 & 0.00 & 0.00 & 0.00  & 0.00\\ \hline  %
Measurement    & Use Thermometer                             & 0.00 & 0.10 & 0.06 & 0.01 & 0.04 & 0.04 & \textbf{0.19}\\ %
Measurement    & Measuring Boiling Point (known)             & 0.00 & 0.08 & \textbf{0.11} & 0.01 & 0.01 & 0.02 & \textbf{0.11}\\ %
Measurement    & Measuring Boiling Point (unknown)           & 0.00 & 0.06 & 0.04 & 0.01 & 0.01 & 0.02 & \textbf{0.10}\\ \hline %
\rowcolor[HTML]{f1f1f1} 
Electricity    & Create a circuit                            & 0.01 & \textbf{0.13} & 0.07 & 0.05 & 0.03 & 0.07 & \textbf{0.13}\\ %
\rowcolor[HTML]{f1f1f1} 
Electricity    & Renewable vs Non-renewable Energy           & 0.01 & 0.10 & 0.04 & 0.07 & 0.02 & 0.05 & \textbf{0.11}\\ %
\rowcolor[HTML]{f1f1f1} 
Electricity    & Test Conductivity (known)                   & 0.01 & 0.07 & 0.04 & 0.02 & 0.05 & 0.05 &\textbf{0.09} \\ %
\rowcolor[HTML]{f1f1f1} 
Electricity    & Test Conductivity (unknown)                 & 0.00 & \textbf{0.20} & 0.04 & 0.02 & 0.04 & 0.05 &0.14 \\ \hline %
Classification & Find a living thing                         & 0.03 & 0.26 & 0.18 & 0.10 & \textbf{0.29} & 0.16  & 0.25\\ %
Classification & Find a non-living thing                     & 0.63 & 0.56 & 0.44 & 0.54 & 0.19 & 0.17  & \textbf{0.94}\\ %
Classification & Find a plant                                & 0.01 & 0.19 & 0.16 & 0.10 & 0.17 & 0.19  & \textbf{0.25}\\ %
Classification & Find an animal                              & 0.01 & 0.19 & 0.15 & 0.08 & 0.21 & 0.19  & \textbf{0.25}\\ \hline %
\rowcolor[HTML]{f1f1f1} 
Biology        & Grow a plant                                & 0.07 & 0.09 & 0.06 & 0.02 & 0.08 & 0.03  & \textbf{0.12}\\ %
\rowcolor[HTML]{f1f1f1} 
Biology        & Grow a fruit                                & 0.02 & \textbf{0.16} & 0.11 & 0.04 & 0.03 & 0.05  & 0.09\\ \hline %
 Chemistry      & Mixing (generic)                            & 0.01 & 0.20 & 0.17 & 0.03 & 0.06 & 0.10 & \textbf{0.22} \\ %
 Chemistry      & Mixing paints (secondary colours)           & 0.01 & 0.29 & 0.19 & 0.06 & 0.16 & 0.20 & \textbf{0.30} \\ %
 Chemistry      & Mixing paints (tertiary colours)            & 0.00 & 0.11 & 0.04 & 0.03 & 0.05 & 0.07 & \textbf{0.14} \\ \hline %
 \rowcolor[HTML]{f1f1f1} 
Biology        & Identify longest-lived animal               & 0.02 & \textbf{0.48} & 0.43 & 0.06 & 0.26 & 0.20 &\textbf{0.48}\\ %
\rowcolor[HTML]{f1f1f1} 
 Biology        & Identify shortest-lived animal              & 0.03 & \textbf{0.47} & 0.32 & 0.10 & 0.14 & 0.16 &0.35\\ %
\rowcolor[HTML]{f1f1f1} 
 Biology        & Identify longest-then-shortest-lived animal & 0.01 & \textbf{0.31} & 0.23 & 0.04 & 0.02 & 0.20 & \textbf{0.31}\\ \hline %
 Biology        & Identify life stages (plant)                & 0.00 & 0.09 & 0.05 & 0.04 & 0.04 & 0.02 & \textbf{0.21}\\ %
Biology        & Identify life stages (animal)               & 0.00 & 0.10 & 0.10 & 0.00 & 0.02 & 0.07 & \textbf{0.18}\\ \hline %
\rowcolor[HTML]{f1f1f1} 
 Forces         & Inclined Planes (determine angle)           & 0.01 & \textbf{0.13} & 0.04 & 0.00 & 0.05 & 0.04 & 0.12\\ %
\rowcolor[HTML]{f1f1f1} 
 Forces         & Friction (known surfaces)                   & 0.00 & \textbf{0.13} & 0.04 & 0.03 & 0.05 & 0.04 & 0.09\\ %
\rowcolor[HTML]{f1f1f1} 
 Forces         & Friction (unknown surfaces)                 & 0.01 & 0.13 & 0.04 & 0.02 & 0.04 & 0.04 & \textbf{0.27}\\ \hline %
 Biology        & Mendelian Genetics (known plants)           & 0.01 & \textbf{0.19} & 0.11 & 0.02 & 0.06 & 0.06 & 0.18 \\ %
Biology        & Mendelian Genetics (unknown plants)         & 0.01 & \textbf{0.17} & 0.11 & 0.02 & 0.13 & 0.05 & 0.16\\ %
\bottomrule
\end{tabular}
\caption{\footnotesize Performance comparison of STARLING against other GRU-based and LLM-based agents on test variations from ScienceWorld. \footnotesize $^{*}$ indicates the reliance on a valid action list during evaluation. 
}%
\label{tab:scienceworld-performance} 
\end{table*}
\subsubsection{ZORK1 Environment}
Zork1 is a human-made interactive fictional game environment and one of the earliest known text-based games created based on the underworld characters with dark themes and characters such as dungeon, grue, elvish sword, etc. Zork1 is one of the 33 interactive games released as a part of the Jericho game suite. Unlike TWC and ScienceWorld, Zork1 includes a diverse set of locations (over $200$ locations), larger action space, sparser rewards, and longer trajectories, making it a challenging environment. 

In order to evaluate the effect of pre-training in the Zork1 environment, we simplified the game with "killing troll" as a final goal \cite{zahavy2018learn}. The agent needs to find the lantern and sword from the house, locate the hidden passageway to the underworld, light the lantern, and kill the troll. Without the lantern and sword, the agent entering the troll room reaches the failure state with negative rewards. In addition to these intermediate rewards, the game includes additional rewards when the agent collects a jewel-encrusted golden egg from the tall tree in the forest (+5 score) and a painting from the art gallery in the house (+4 score). We train the agents on $100$ episodes with a maximum step of $100$ steps per episode.

Figure \ref{fig:zork_results} (left) shows the performance of both the STARLING and vanilla TBRL on the Zork1 environment. As in TWC, we can see that the pre-training step boosts the performance of STARLING in the first few episodes compared to the vanilla TBRL agent. As in ScienceWorld, STARLING successfully avoids the failure state compared to the vanilla TBRL agent.

 \subsubsection{ScienceWorld Environment}

ScienceWorld environment evaluates the science reasoning abilities of the TBRL agents. It consists of several tasks from topics such as change of state, biology, classification, etc.  We choose all the 30 science-domain tasks from the themes.
Each of these tasks contains $10-1400$ variations of the game and are split into $50\%$ training, $25\%$ for evaluation set, and $25\%$ for test set. We train the STARLING agent with $100k$ maximum steps on a single environment (with a maximum of $100$ steps per game play)\footnote{Unlike in the "number of moves taken" metric in the pre-training results, the number of moves taken in the ScienceWorld measures how long the agent survived without reaching the failure state. An agent may reach a failure state if it takes an action that results in abruptly ending the game such as pouring water on the floor for the \textit{boiling} task, etc}.

Figure \ref{fig:sw_training_curves} shows the training curves for both the scores received and moves taken on the 4 tasks. We can see that STARLING outperforms vanilla TBRL on all the tasks. We notice that the pre-training steps improved the performance of STARLING in the first few episodes of the classification task (find a living thing). On the other hand, pre-training games such as boiling water, cooking pasta, planting a tree, etc. may have influenced the performance of STARLING in the later episodes of tasks: change of state - boiling and grow a fruit, by adapting the learned skills during pre-training to the target environment. %

Table \ref{tab:scienceworld-performance} shows the performance comparison of GRU-based and LLM-based models against STARLING on the test variations of the ScienceWorld. We compare the STARLING against the state-of-the-art LLM-based models Behavior cloning (BC), Text Decision Transformers (TDT), CALM-GPT2, etc \cite{wang2023behavior,ammanabrolugraph,yao2020keep}.   We notice that our pretraining strategy has generally improved the performance of STARLING in the majority of the tasks. We outperform both GRU-based and LLM-based models. It is worth noting that both BC and TDT use T5-base (11 billion parameters) initialized with Macaw, CALM uses GPT2 with 131 million parameters, whereas, GRU-based STARLING uses approx 200K parameters. 

\subsection{Discussion}
Our experiments on three text-based game environments show that pre-training these agents using LLM-generated games as auxiliary tasks generally boosts the performance of the agent. We notice that most of the performance gain achieved in the first few episodes of the gameplay on some game instances (e.g., training curves in ScienceWorld tasks find a living thing, TWC Easy, and Hard, Zork1), whereas, STARLING adapts the basic skills learned during the pre-training to the target environment to improve the final scores (e.g., training curves in ScienceWorld tasks: change of state - boiling and grow a fruit). We notice that STARLING avoids the failure states better than vanilla TBRL as can be seen in Zork1 and ScienceWorld. Similarly, STARLING tends to choose valid actions from the action space more effectively than vanilla TBRL (for example, see Figure \ref{fig:sw_sample} supplementary for the sample trajectories from ScienceWorld taken by vanilla TBRL and STARLING within the first few episodes for the task: changes of state - boiling). 

Since these pre-training games lack navigational complexity that elicits skills such as planning, we observe that STARLING tends to suffer in games that require navigational skills (e.g., example, in ScienceWorld task find a non-living thing, TWC hard difficulty and Zork1). Since the pre-training games involve fewer sequences of actions (short trajectories) to collect the reward and reach the final goal, STARLING struggles when the target environment has longer trajectories to reach the goal. On the other hand, STARLING tends to collect bonus scores in Zork1 that are reachable within fewer steps instead of just chasing the larger rewarded states (e.g., see Figure \ref{fig:zork_results} right). 
\section{Related Work}
\subsection{Text-based Games}

Text-based games provide a challenging benchmark for RL agents interacting with the environment in natural language. The common challenges for text-based RL are partial observability, combinatorial action space, sparse rewards, long-horizon planning, etc. To circumvent some of these challenges, it is typical that the games are often manually curated to evaluate a specific set of skills such as commonsense reasoning \cite{murugesan2021text}, knowledge graphs \cite{ammanabrolugraph,murugesan2021efficient}, exploration strategies \cite{cote2019textworld}, etc. Environments such as TextWorld Commonsense \cite{murugesan2021text} measure simple commonsense reasoning based on the one-hop relationship between a pair of everyday objects such (\textit{apple} and \textit{refrigerator}, \textit{dirty towel} and \textit{laundry basket}), but lacks diversity and complexity to learn a general set of skills. 

Environments such as ScienceWorld \cite{wang2022scienceworld} are often domain-specific environments that require domain knowledge to perform well in these environments. Jericho \cite{hausknecht2020interactive}, on the other hand, includes human-generated games that require a complex set of skills to show any progress in the gameplay. These environments are manually created by humans with very limited automation in the variations of game generation by replacing similar or related objects, changing the layout/orientation of the environment, etc.  Unlike these environments, STARLING provides an approach to leverage the skill generation capability of LLMs \cite{huang2022language} to automatically generate text-based games based on the input game ideas with minimal human supervision. These games are generated automatically by requesting a specific set of game-related facts from LLM in a slot-filling style text generation \cite{rakotonirina2022can}.

\subsection{Self-supervised RL}
Self-supervised RL has been a popular topic in vision-based RL and robotic environments \cite{sekar2020planning,li2022does}.
To the best of our knowledge, we are the first to utilize LLM to generate the games to train text-based RL agents\footnote{Several LLM-based interactive fictional environments for entertainment purposes exist \cite{AIDungeon}.}. Previous works have utilized LLM for action generation \cite{yao2020keep}, play interactive fictional game \cite{tsai2023can}, build a world model \cite{ammanabrolu2020avoid,wang2023bytesized32}, etc. These works showed that using LLM to learn the underlying representation of text in the environment does not necessarily improve the performance  \citep{wang2022scienceworld}. It is shown that novel exploration strategies and efficient RL algorithms along with the learning model similar to the one in Figure \ref{fig:tbrl} outperform all the other LLM-based agents \cite{tuyls2021multi}. On the other hand, generalist agents have been recently explored to generalize across multiple environments \cite{reed2022generalist}, but the performance of these agents on a diverse set of environments is less convincing \cite{cobbe2019quantifying}.

\section{Conclusion}
In this paper, we proposed a novel self-supervised training for a text-based reinforcement learning agent, STARLING, with the help of the generalized skill generation capability of large language models like GPT3. 
We generated a set of text-based games that require agents to learn basic skills such as cooking pasta, boiling water, etc., and utilize sequential decision-making over the modality of text. The proposed STARLING uses the GPT3 pre-trained language model to automatically generate these games. This approach can be used to create additional games or adapted to build games for new domains with minimal human intervention. We showed that the STARLING agent pre-trained on the games generated by LLM outperforms vanilla TBRL.  We evaluated STARLING on three environments: ScienceWorld, TWC, and Zork1. In all these environments, STARLING showed enhanced skills in the target environment. 

\newpage
\section{Limitations}
Human participants were volunteers from a local High School that agreed to participate in this study. This may have introduced a bias into the human participant data since all participants were high school educated, from one geographic region, between the ages of 15 and 18, and volunteers. Many of these participants complete homework assignments and assessments often which may make their reasoning skills better than potential participants outside of school. In the future, testing human participants from various geographic locations, age groups, and levels of education may reduce bias. 
The STARLING currently requires human intervention and/or the Game Validator (from the glulx compiler) to build functioning games. We will continue to work on building an end-to-end version of the STARLING, that can take a game idea and turn it into an interactive fiction game without any human intervention. This would speed up development time so a larger set of games can be created. 

Large language models such as ChatGPT have been developed recently with the ability to interact with users in a manner conversationally similar to the interactions found in interactive fiction games. From our experimentation, ChatGPT struggles to keep track of the states of all in-game objects and the pre-conditions necessary to use those actions (e.g. ChatGPT does not always require the player to turn on the stove before using it) as well as Inform7-based games. In addition, it suffers from small factual errors, and is hard to reproduce the same result, through this could also be seen as a benefit. Despite these challenges exploring the use of models such as ChatGPT to interact with agents shows promise in the future. 

\section{Ethical Impact}

We asked the human participants to play the games generated by STARLING to evaluate the game's complexity and clarity. 
After receiving IRB approval from a local High School and informed consent from each of the $48$ human participants, we asked the participants to play five randomly assigned games via \url{iplayif.com}, an online interactive fiction game player. Players received the goal of the game and the list of admissible actions. We collected the number of steps that each player took and the score received for each game via Google form. We did not collect any personal information or personally identifiable information as a part of this study.

Since we use large language models such as GPT3 to generate a set of text-based games, the bias and other fairness/ethical concerns that come with the LLM may unintentionally transfer to the pre-trained agent. Additional mitigation steps may be required to filter harmful contents from the generated response.

\section{Reproducibility}
As an effort to encourage further research in self-supervised text-based RL, we plan to release the $100$ games generated as a part of this paper, the source code for STARLING, script to generate game-related files based on a set of game ideas and LLM (including game templates and metadata) as an open-source project.
\bibliography{starling}
\appendix
\input{supp}

\end{document}

%% file: supp.tex
\begin{figure*}[ht]
    \centering
    \includegraphics[width = 0.45\linewidth]{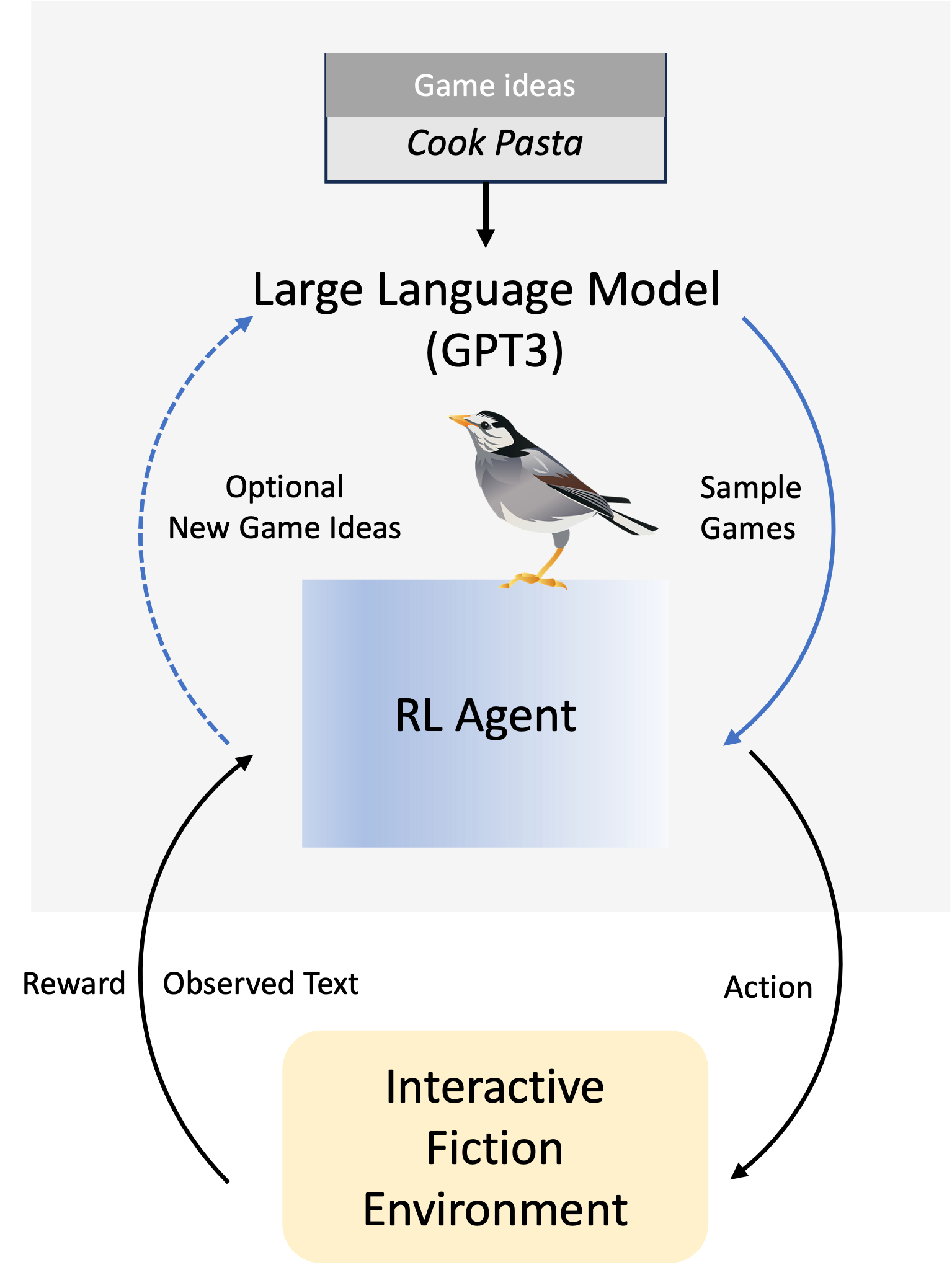}
    \includegraphics[width = 0.42\linewidth]{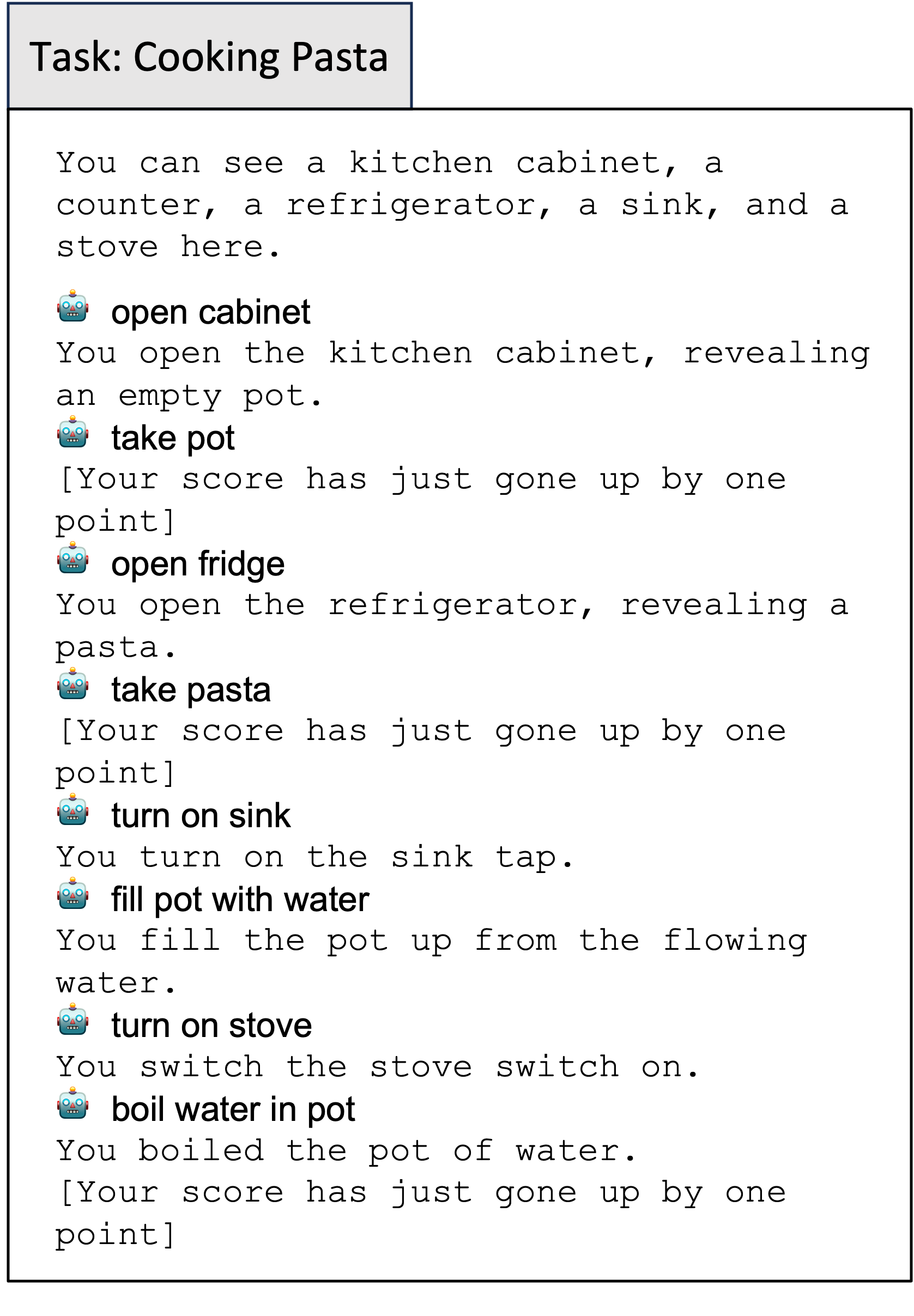}
    \caption{(Left) shows the overview of the proposed self-supervised text-based reinforcement learning with large language model. (Right) shows the sample text-only agent play through of cooking pasta game. Players must use the boil skill at the correct time to be successful.}
    \label{fig:architecture_diagram}
\end{figure*}
\begin{figure*}[ht]
    \centering
    \includegraphics[scale=0.5]{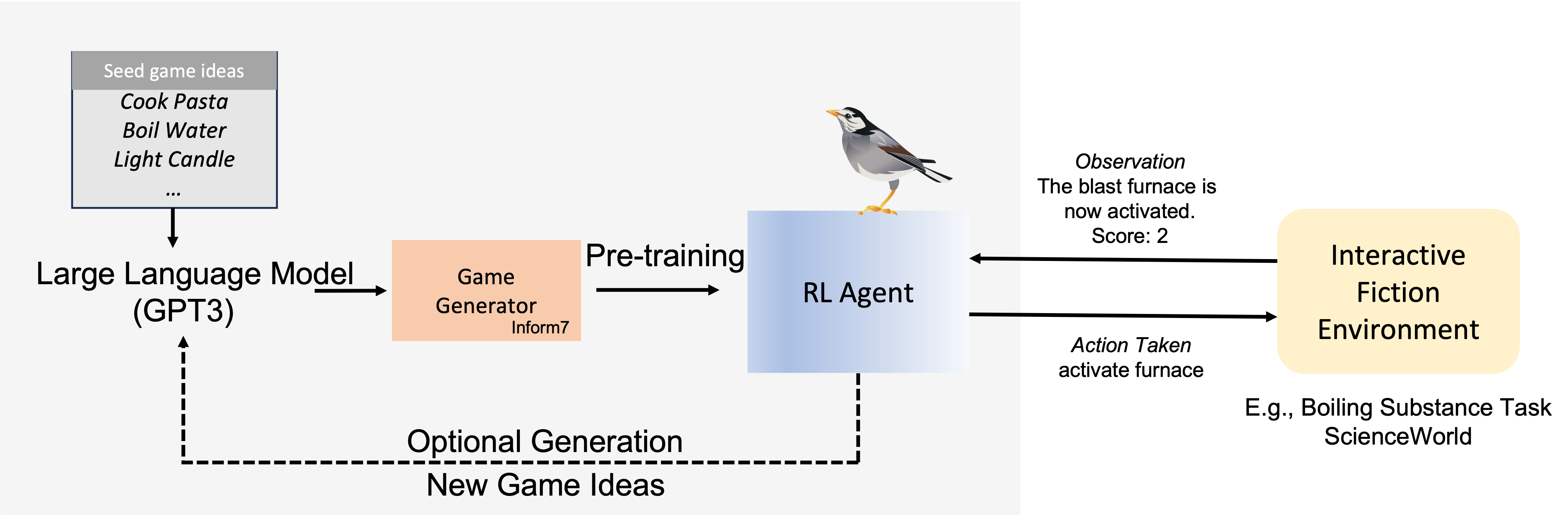}
    \caption{Architecture diagram for Self-supervised Text-based Reinforcement Learning using LLM (STARLING).}
    \label{fig:arch_full}
\end{figure*}

\section{Additional Details}
In order to generate games that require composing previously learned skills, we take inspiration from household chores, cooking, and maintenance tasks. We generate $100$ game ideas and use the STARLING to generate a set of $100$ games. We choose the game ideas carefully for the learning agent to utilize similar skills (ex. baking, mixing, spreading, using a hammer, etc) in new situations therefore forcing the agent to generalize skills and compose them with other skills. For example, while cooking pasta, an agent must learn how to boil water which is a skill that can be applied for a related game idea "brewing tea".

LLMs such as GPT-3 are prone to making factual and grammatical errors, in addition to violating the specified format. We check for any errors in the generated game(s) using a Game Validator as a part of \textit{glulx} compiler which uses depth first search (DFS) to explore all the possible trajectories in the game. To correct for minor errors and inconsistencies in each game, information from GPT-3 can also be optionally verified by the human authors in the JSON file. We found that, in cases when the created game has errors, restarting the game generation a few times usually results in a playable game.

\subsection{From GPT3 output to game JSON}

We extract the information from GPT-3 using Python simple regular expression rules by first splitting the output into three sections: task sequence (ex. Open cabinet, take pot), objects (ex. Pot), and actions (ex. Fill pot with water). We add the task sequence to the list of admissible actions the player could execute within the game. We store the objects internally with a type, a location, a name, and a set of properties. We further split the actions section into default actions and custom actions which are actions native and non-native to inform7 respectively. Similar to the objects, we store each custom action internally with a name, a declaration, a definition, a set of constraints, a set of prerequisites, and a set of post-requisites. 

\subsection{Inform7 Code Generator}
We wrote a simple script based on the JSON structure and the action templates to generate the Inform7 Code for a given game idea. This script along with the other code \& data will be shared to the public in the near future.

\subsection{Modifying TextWorld Gym for STARLING}
OpenAI Gym is a general reinforcement learning framework that acts as an interface between RL agents and Inform7-based STARLING game engine \cite{1606.01540}. Gym connects environments with agents by using a monitor to keep track of every step, state of the game, the final score of agents, and the sample complexity or the amount of time an agent takes to learn. 
Most default environments in Gym support a continuous or discrete action space although interactive fiction games require combinatorial action spaces in natural language \cite{hausknecht2020interactive}. The TextWorld Gym customized the OpenAI Gym for interactive fiction games.  In this work, we repurpose the custom Gym environment created for TextWorld environment with Inform7 object and action types. 

TextWorld’s Gym environment only supports TextWorld-generated games which includes a Glulx compiled game file and a TextWorld-generated JSON file with game metadata defined in proprietary TextWorld classes. This restricted our ability to create games with objects and actions previously undefined in TextWorld environment. These objects and actions must be defined according to TextWorld's grammar and logic rules. This is a time consuming process and is prone to many errors. The goal of the STARLING Game Generator is to allow users to automate the game creation using LLM, and most importantly, create games without learning a new programming language or familiarizing themselves with any grammar rules. 

To get rid of these restrictions, an entirely new wrapper was created which acted as an interface between the game engine and TextWorld Gym environment. This wrapper ensures that the user to freely define any object or action type and  the environment works with any Glulx compiled game file without dependence on the TextWorld-generated metadata to track the state of objects throughout the game. The wrapper does this by parsing the observation state returned by game engine after every step to generate certain data-points like admissible commands, current score, last action, number of steps taken and inventory required by the TextWorld Gym environment.

\section{Human Participants}
\label{sec:human}
Humans are considered to have exemplary compositional skill learning so comparing their performance to pre-trained agent's performance is valuable to validate generated games's difficulty and effectiveness as a pre-training task. After receiving IRB approval from a local High School and informed consent from each of the $48$ human participants, we asked the participants to play five randomly assigned games via \url{iplayif.com}, an online interactive fiction game player. Players received the goal of the game and the list of admissible actions. We collected the number of steps that each player took and the score received for each game via Google form. \footnote{No personal information was collected as a part of this study.}

\section{Pretraining Game Statistics}
\begin{table}[h]
    \centering
    \resizebox{0.4\textwidth}{!}{%
    \begin{tabular}{l|c}
    \hline
    \multicolumn{2}{c}{\textbf{Game-specific Statistics}}  \\ \hline
    \textit{Min. \# Actions}      & 7.36 $\pm$ 2.53          \\
    \textit{Avg. Rewards across games}       & 4.08 $\pm$ 1.57 \\
    \textit{Num. Skills per game}           & 2 $\pm$ 1          \\ 
    \hline
    \end{tabular}
    }
    \label{tab:game_specifics}
\end{table}

\begin{figure*}[h]
    \centering
    \includegraphics[width = 0.95\linewidth]{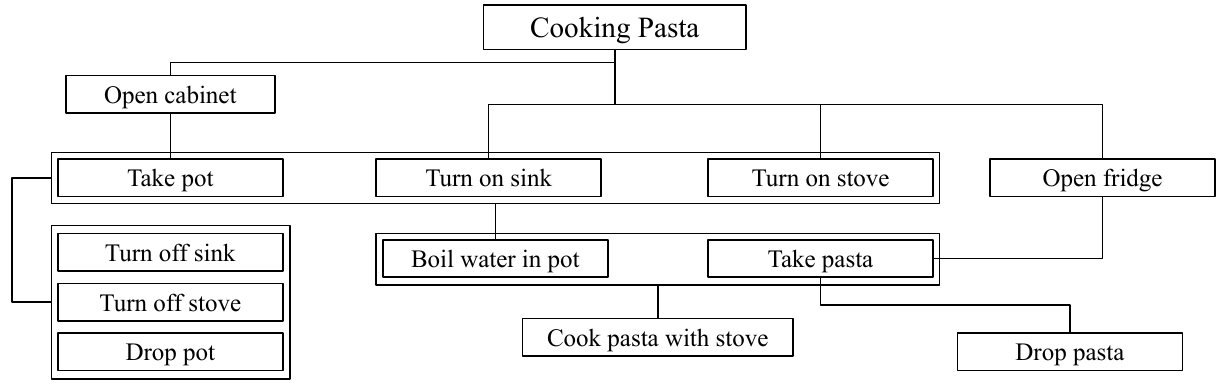}
    \caption{Composition of skills needed to complete the game "\textit{Cooking Pasta}" as Flow diagram. A line between two skills represent that one skill needs to be executed before executing the other one (E.g., \textit{Open cabinet} $\longrightarrow$ \textit{Take pot}). A box with multiple skills represent that skills within the box can be executed in any order (E.g., \textit{Boil water in pot} $\parallel$ \textit{Take pasta}).}
    \label{fig:flow_diagram}
\end{figure*}

\begin{figure*}[h]
  \centering
     \includegraphics[width=0.7\textwidth]{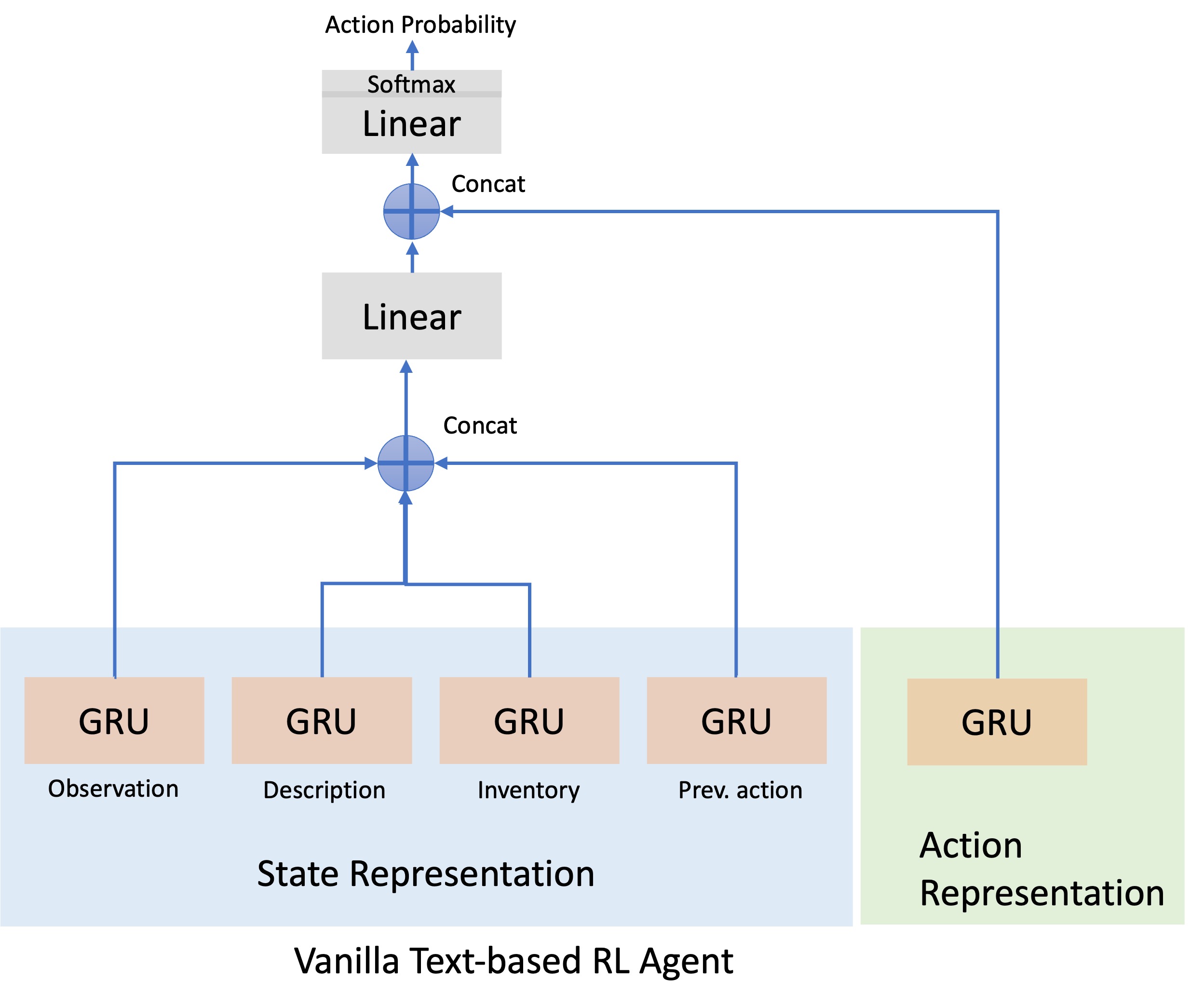}
\    \caption{Vanilla Text-based RL agent used in this paper. }
    \label{fig:tbrl}
\end{figure*}

\begin{figure*}[h]
  \centering
     \includegraphics[width=0.7\textwidth]{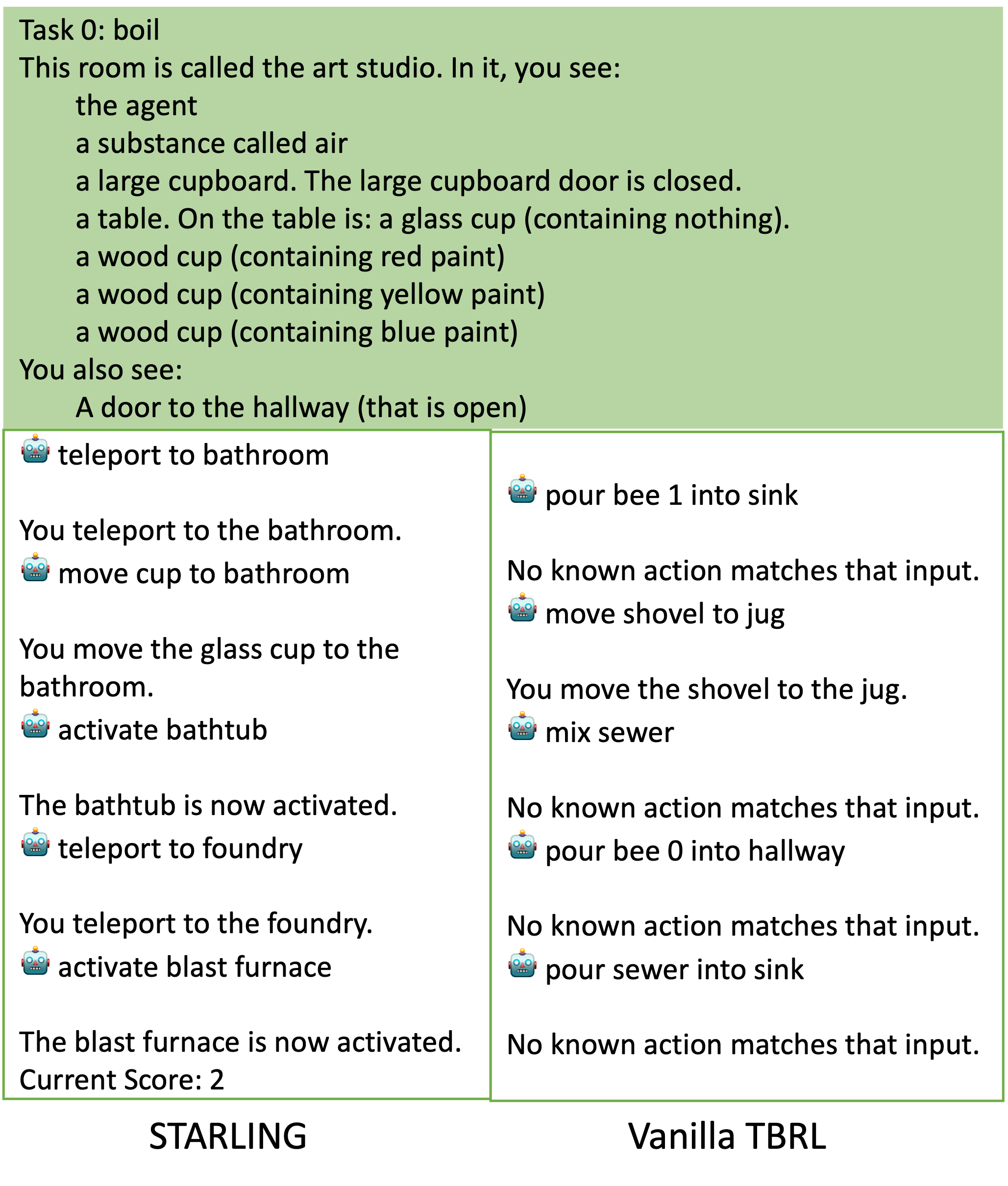}
\    \caption{Sample Trajectories taken by STARLING and Vanilla TBRL for the task: change of state (boiling) task in ScienceWorld.}
    \label{fig:sw_sample}
\end{figure*}

\begin{figure*}[!h]
     \centering
     \includegraphics[width=0.75\linewidth]{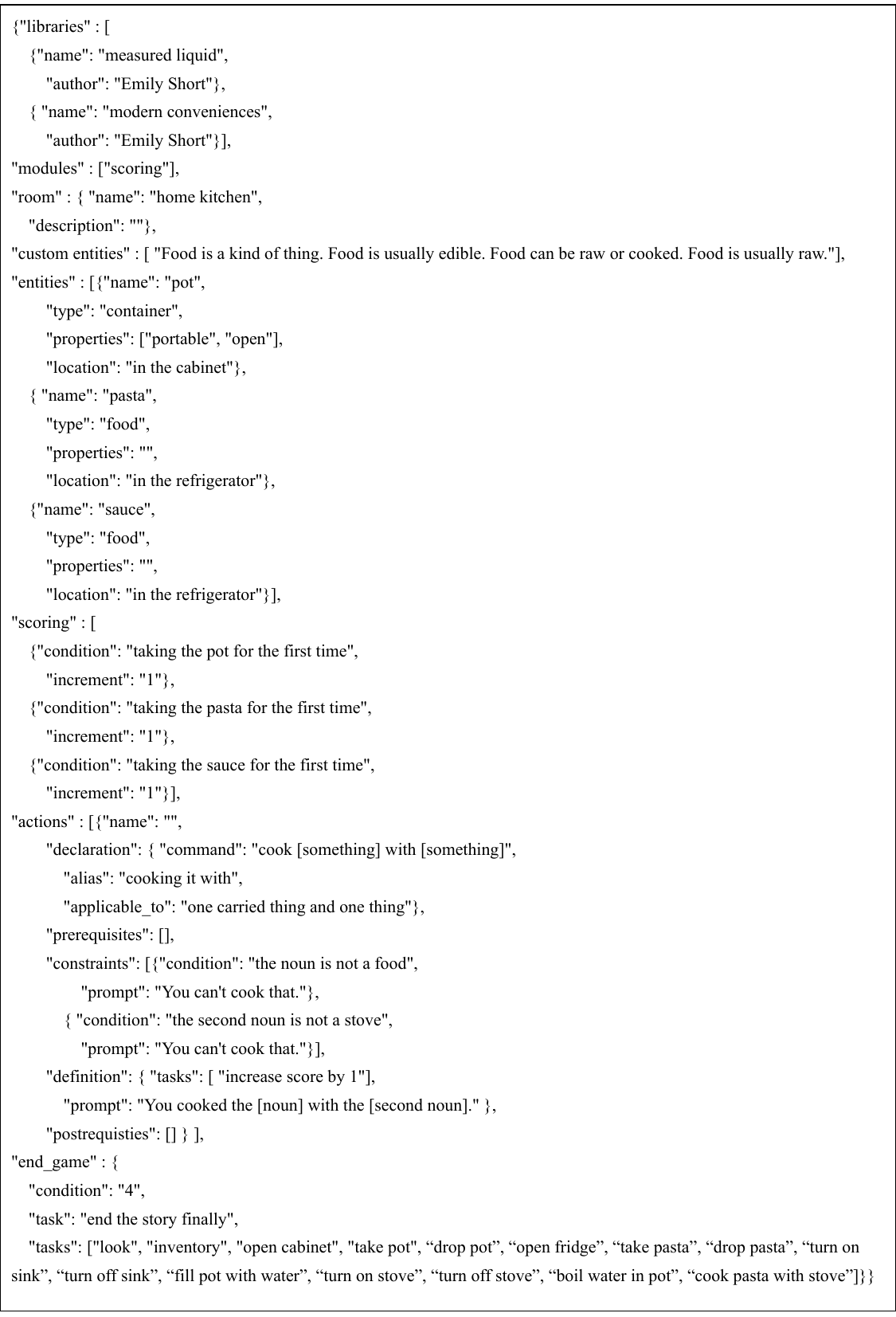}
    \caption{Example JSON file produced for cooking pasta game idea. The libraries, modules, and room sections were part of the setup, the custom entities and entities sections correspond to object creation, the actions correspond to the custom actions, and the scoring and end game correspond to the rewards sections of each game. The entities section describes names, types, and properties of entities present in the game. The actions section defines custom actions including their declaration, alias, and constraints not part of Inform7 by default. The end-game section defines the maximum score and the list of admissible actions that the user can take. }
    \label{fig:json_file}
\end{figure*}